\documentclass{article} 
\usepackage{arxiv,times}

\usepackage{amsmath,amsfonts,bm}

\def\eqref#1{equation~\ref{#1}}

\def\1{\bm{1}}

\DeclareMathAlphabet{\mathsfit}{\encodingdefault}{\sfdefault}{m}{sl}
\SetMathAlphabet{\mathsfit}{bold}{\encodingdefault}{\sfdefault}{bx}{n}

\usepackage[T1]{fontenc}

\usepackage[utf8]{inputenc}

\usepackage{microtype}

\usepackage{inconsolata}

\usepackage{algorithm}
\usepackage{algpseudocode}
\usepackage{amsfonts}
\usepackage{amsmath}
\usepackage{amssymb}
\usepackage{array}
\usepackage{bm}
\usepackage{bbm}
\usepackage{booktabs}
\usepackage{colortbl}
\usepackage{dsfont}
\usepackage{enumitem}
\usepackage{graphicx}
\usepackage{hyperref}
\usepackage{latexsym}
\usepackage{longtable}
\usepackage{mathtools}
\usepackage{multirow}
\usepackage{multicol}
\usepackage{nicefrac}
\usepackage{tabularx}
\usepackage{times}
\usepackage{pgf}
\usepackage{pifont}
\usepackage{tcolorbox}
\usepackage{tikz}
\usepackage{url}
\usepackage{wrapfig}
\usepackage{xcolor}
\usepackage{xspace}

\title{It Takes Workflows to Evolve Better Workflows}

\author{
\makebox[\textwidth][c]{Xuehang Guo\textsuperscript{1}, Haoyu Wang\textsuperscript{2}, Haifeng Chen\textsuperscript{2}, Yangyi Chen\textsuperscript{3}, Zhenhailong Wang\textsuperscript{3}\thanks{Equal co-mentorship. \\ Correspondence: Xuehang Guo \texttt{<xguo15@wm.edu>}, Zhenhailong Wang \texttt{<wangz3@illinois.edu>}, Qingyun Wang \texttt{<qwang16@wm.edu>}}, Qingyun Wang\textsuperscript{1*}} \\
\makebox[\textwidth][c]{\textsuperscript{1}William \& Mary \quad \textsuperscript{2}NEC Corporation of America \quad \textsuperscript{3}University of Illinois Urbana-Champaign}
}

\definecolor{ourscolor}{HTML}{5E82AE}

\definecolor{mycitecolor}{HTML}{6FA9DF}

\definecolor{bboxedge}{HTML}{95B3D0}
\definecolor{bboxface}{HTML}{E7EDF4}
\definecolor{bboxtag}{HTML}{4C7CA4}

\definecolor{notunecolor}{HTML}{888888}

\definecolor{sandboxcolor}{HTML}{EF8537}
\definecolor{teachergreen}{HTML}{5CB86F}

\definecolor{BASELINE_BG}{HTML}{E6E6E6}
\definecolor{STUDENT_BG}{HTML}{E0EFF8}
\definecolor{STUDENT_AUG_BG}{HTML}{E0EFF8}
\definecolor{TEACHER_BG}{HTML}{D6F1E0}
\definecolor{TEACHER_AUG_BG}{HTML}{D6F1E0}

\definecolor{gray_bg}{HTML}{F2F2F2}
\definecolor{gray_text}{HTML}{7D7D7D}
\definecolor{blue_bg}{HTML}{E7EDF4}
\definecolor{blue_text}{HTML}{3D6383}
\definecolor{tea_bg}{HTML}{E5F1EC}
\definecolor{tea_text}{HTML}{598C7C}
\definecolor{rose_bg}{HTML}{FAECEC}
\definecolor{rose_text}{HTML}{B27070}

\definecolor{tea_delta}{HTML}{66AD8F}

\newcommand{\dcell}[2]{\cellcolor{tea_delta!#1}#2}

\definecolor{deltagreen}{HTML}{B1D3A8}
\definecolor{deltared}{HTML}{EA8C8C}
\definecolor{deltaredbg}{HTML}{F4AAAA}
\colorlet{deltacellcolor}{white}

\definecolor{checkgreen}{HTML}{31B331}
\definecolor{crossred}{HTML}{E6716E}
\definecolor{checkgreen_bg}{HTML}{DFF6DF}
\definecolor{crossred_bg}{HTML}{FDE6E6}

\hypersetup{
    colorlinks=true,
    linkcolor=mycitecolor,
    citecolor=mycitecolor,
    urlcolor=mycitecolor
}

\newcommand{\ours}{\textcolor{ourscolor}{\textbf{\textsc{FloWright}}}\xspace}
\newcommand{\oursdata}{\textcolor{ourscolor}{\textbf{\textsc{DataWright}}}\xspace}

\tcbuselibrary{skins,breakable}

\newtcolorbox{hlbox}[1]{
  enhanced, breakable,
  colback=bboxface, colframe=bboxface, boxrule=0pt, arc=6pt,
  left=8pt, right=8pt, top=12pt, bottom=6pt,
  title=#1,
  fonttitle=\footnotesize\bfseries, coltitle=white,
  attach boxed title to top left={xshift=10pt, yshift=-\tcboxedtitleheight/2},
  boxed title style={colback=bboxtag, colframe=bboxtag, boxrule=0pt, arc=3pt,
                     left=6pt, right=6pt, top=2pt, bottom=2pt},
}

\newcommand{\circnum}[1]{%
  \begin{tikzpicture}[baseline=-0.6ex]
    \node[circle, fill=bboxedge, inner sep=0.6pt,
          minimum size=6pt, text=white,
          font=\bfseries\small] {#1};
  \end{tikzpicture}%
}

\newlist{oursenumerate}{enumerate}{1}

\setlist[oursenumerate,1]{
  label=\protect\circnum{\arabic*},
  leftmargin=1.5em,
  itemsep=0pt,
  topsep=0pt,
  partopsep=0pt,
  parsep=0pt,
}

\algrenewcommand{\algorithmiccomment}[1]{\hfill\textcolor{gray}{\(\triangleright\) #1}}

\newcommand{\one}{\mathds{1}}

\newcommand{\applydeltacellcolor}[1]{%
  \pgfmathsetmacro{\myintensity}{int(min(100, abs(#1) * 6))}%
  \ifdim#1pt>0pt%
    \edef\tmpcolor{deltagreen!\myintensity!white}%
  \else%
    \edef\tmpcolor{deltaredbg!\myintensity!white}%
  \fi%
  \expandafter\cellcolor\expandafter{\tmpcolor}%
}

\newcommand{\myul}[1]{\leavevmode\vtop{\vbox{\hbox{#1}}\hrule height 0.4pt}}

\newcommand{\testvaldeltabgpct}[2]{%
  \pgfmathsetmacro{\deltares}{#2-#1}%
  \pgfmathtruncatemacro{\valwhole}{int(#2)}%
  \pgfmathtruncatemacro{\valdec}{int(round((#2-\valwhole)*100))}%
  \ifnum\valdec<10\edef\valdecstr{0\valdec}\else\edef\valdecstr{\valdec}\fi%
  \ifdim\deltares pt>0pt%
    \pgfmathsetmacro{\deltaint}{int(min(100, abs(\deltares) * 3))}%
    \edef\deltacolor{deltagreen!\deltaint!white}%
    \expandafter\cellcolor\expandafter{\deltacolor}\small\rmfamily\valwhole.\valdecstr%
  \else%
    \pgfmathsetmacro{\deltaint}{int(min(100, abs(\deltares) * 2))}%
    \edef\deltacolor{deltared!\deltaint!white}%
    \expandafter\cellcolor\expandafter{\deltacolor}\small\rmfamily\valwhole.\valdecstr%
  \fi%
}

\newcommand{\testvaldeltabgpctbf}[2]{%
  \pgfmathsetmacro{\deltares}{#2-#1}%
  \pgfmathtruncatemacro{\valwhole}{int(#2)}%
  \pgfmathtruncatemacro{\valdec}{int(round((#2-\valwhole)*100))}%
  \ifnum\valdec<10\edef\valdecstr{0\valdec}\else\edef\valdecstr{\valdec}\fi%
  \ifdim\deltares pt>0pt%
    \pgfmathsetmacro{\deltaint}{int(min(100, abs(\deltares) * 3))}%
    \edef\deltacolor{deltagreen!\deltaint!white}%
    \begingroup\rmfamily\bfseries\small\expandafter\cellcolor\expandafter{\deltacolor}\valwhole.\valdecstr\endgroup%
  \else%
    \pgfmathsetmacro{\deltaint}{int(min(100, abs(\deltares) * 2))}%
    \edef\deltacolor{deltared!\deltaint!white}%
    \begingroup\rmfamily\bfseries\small\expandafter\cellcolor\expandafter{\deltacolor}\valwhole.\valdecstr\endgroup%
  \fi%
}

\newcommand{\testvaldeltabgpctul}[2]{%
  \pgfmathsetmacro{\deltares}{#2-#1}%
  \pgfmathtruncatemacro{\valwhole}{int(#2)}%
  \pgfmathtruncatemacro{\valdec}{int(round((#2-\valwhole)*100))}%
  \ifnum\valdec<10\edef\valdecstr{0\valdec}\else\edef\valdecstr{\valdec}\fi%
  \ifdim\deltares pt>0pt%
    \pgfmathsetmacro{\deltaint}{int(min(100, abs(\deltares) * 3))}%
    \edef\deltacolor{deltagreen!\deltaint!white}%
    \expandafter\cellcolor\expandafter{\deltacolor}\small\rmfamily\myul{\valwhole.\valdecstr}%
  \else%
    \pgfmathsetmacro{\deltaint}{int(min(100, abs(\deltares) * 2))}%
    \edef\deltacolor{deltared!\deltaint!white}%
    \expandafter\cellcolor\expandafter{\deltacolor}\small\rmfamily\myul{\valwhole.\valdecstr}%
  \fi%
}

\newcommand{\testvaldeltabgpctbful}[2]{%
  \pgfmathsetmacro{\deltares}{#2-#1}%
  \pgfmathtruncatemacro{\valwhole}{int(#2)}%
  \pgfmathtruncatemacro{\valdec}{int(round((#2-\valwhole)*100))}%
  \ifnum\valdec<10\edef\valdecstr{0\valdec}\else\edef\valdecstr{\valdec}\fi%
  \ifdim\deltares pt>0pt%
    \pgfmathsetmacro{\deltaint}{int(min(100, abs(\deltares) * 3))}%
    \edef\deltacolor{deltagreen!\deltaint!white}%
    \begingroup\rmfamily\bfseries\small\expandafter\cellcolor\expandafter{\deltacolor}\myul{\valwhole.\valdecstr}\endgroup%
  \else%
    \pgfmathsetmacro{\deltaint}{int(min(100, abs(\deltares) * 2))}%
    \edef\deltacolor{deltared!\deltaint!white}%
    \begingroup\rmfamily\bfseries\small\expandafter\cellcolor\expandafter{\deltacolor}\myul{\valwhole.\valdecstr}\endgroup%
  \fi%
}

\newcommand{\trainvaldeltabgpct}[2]{%
  \pgfmathsetmacro{\deltares}{#2-#1}%
  \pgfmathtruncatemacro{\valwhole}{int(#2)}%
  \pgfmathtruncatemacro{\valdec}{int(round((#2-\valwhole)*100))}%
  \ifnum\valdec<10\edef\valdecstr{0\valdec}\else\edef\valdecstr{\valdec}\fi%
  \ifdim\deltares pt>0pt%
    \pgfmathsetmacro{\deltaint}{int(min(100, abs(\deltares) * 20))}%
    \edef\deltacolor{deltagreen!\deltaint!white}%
    \expandafter\cellcolor\expandafter{\deltacolor}\small\rmfamily\valwhole.\valdecstr%
  \else%
    \pgfmathsetmacro{\deltaint}{int(min(100, abs(\deltares) * 6))}%
    \edef\deltacolor{deltared!\deltaint!white}%
    \expandafter\cellcolor\expandafter{\deltacolor}\small\rmfamily\valwhole.\valdecstr%
  \fi%
}

\newcommand{\trainvaldeltabgpctbf}[2]{%
  \pgfmathsetmacro{\deltares}{#2-#1}%
  \pgfmathtruncatemacro{\valwhole}{int(#2)}%
  \pgfmathtruncatemacro{\valdec}{int(round((#2-\valwhole)*100))}%
  \ifnum\valdec<10\edef\valdecstr{0\valdec}\else\edef\valdecstr{\valdec}\fi%
  \ifdim\deltares pt>0pt%
    \pgfmathsetmacro{\deltaint}{int(min(100, abs(\deltares) * 20))}%
    \edef\deltacolor{deltagreen!\deltaint!white}%
    \begingroup\rmfamily\bfseries\small\expandafter\cellcolor\expandafter{\deltacolor}\valwhole.\valdecstr\endgroup%
  \else%
    \pgfmathsetmacro{\deltaint}{int(min(100, abs(\deltares) * 6))}%
    \edef\deltacolor{deltared!\deltaint!white}%
    \begingroup\rmfamily\bfseries\small\expandafter\cellcolor\expandafter{\deltacolor}\valwhole.\valdecstr\endgroup%
  \fi%
}

\newcommand{\trainvaldeltabgpctul}[2]{%
  \pgfmathsetmacro{\deltares}{#2-#1}%
  \pgfmathtruncatemacro{\valwhole}{int(#2)}%
  \pgfmathtruncatemacro{\valdec}{int(round((#2-\valwhole)*100))}%
  \ifnum\valdec<10\edef\valdecstr{0\valdec}\else\edef\valdecstr{\valdec}\fi%
  \ifdim\deltares pt>0pt%
    \pgfmathsetmacro{\deltaint}{int(min(100, abs(\deltares) * 20))}%
    \edef\deltacolor{deltagreen!\deltaint!white}%
    \expandafter\cellcolor\expandafter{\deltacolor}\small\rmfamily\myul{\valwhole.\valdecstr}%
  \else%
    \pgfmathsetmacro{\deltaint}{int(min(100, abs(\deltares) * 6))}%
    \edef\deltacolor{deltared!\deltaint!white}%
    \expandafter\cellcolor\expandafter{\deltacolor}\small\rmfamily\myul{\valwhole.\valdecstr}%
  \fi%
}

\newcommand{\trainvaldeltabgpctbful}[2]{%
  \pgfmathsetmacro{\deltares}{#2-#1}%
  \pgfmathtruncatemacro{\valwhole}{int(#2)}%
  \pgfmathtruncatemacro{\valdec}{int(round((#2-\valwhole)*100))}%
  \ifnum\valdec<10\edef\valdecstr{0\valdec}\else\edef\valdecstr{\valdec}\fi%
  \ifdim\deltares pt>0pt%
    \pgfmathsetmacro{\deltaint}{int(min(100, abs(\deltares) * 20))}%
    \edef\deltacolor{deltagreen!\deltaint!white}%
    \begingroup\rmfamily\bfseries\small\expandafter\cellcolor\expandafter{\deltacolor}\myul{\valwhole.\valdecstr}\endgroup%
  \else%
    \pgfmathsetmacro{\deltaint}{int(min(100, abs(\deltares) * 6))}%
    \edef\deltacolor{deltared!\deltaint!white}%
    \begingroup\rmfamily\bfseries\small\expandafter\cellcolor\expandafter{\deltacolor}\myul{\valwhole.\valdecstr}\endgroup%
  \fi%
}

\newcommand{\chartdeltabgpct}[2]{%
  \pgfmathsetmacro{\deltares}{#2-#1}%
  \pgfmathtruncatemacro{\valwhole}{int(#2)}%
  \pgfmathtruncatemacro{\valdec}{int(round((#2-\valwhole)*100))}%
  \ifnum\valdec<10\edef\valdecstr{0\valdec}\else\edef\valdecstr{\valdec}\fi%
  \ifdim\deltares pt>0pt%
    \pgfmathsetmacro{\deltaint}{int(min(100, abs(\deltares) * 1.8))}%
    \edef\deltacolor{deltagreen!\deltaint!white}%
    \expandafter\cellcolor\expandafter{\deltacolor}\small\rmfamily\valwhole.\valdecstr%
  \else%
    \pgfmathsetmacro{\deltaint}{int(min(100, abs(\deltares) * 2))}%
    \edef\deltacolor{deltared!\deltaint!white}%
    \expandafter\cellcolor\expandafter{\deltacolor}\small\rmfamily\valwhole.\valdecstr%
  \fi%
}

\newcommand{\chartdeltabgpctbf}[2]{%
  \pgfmathsetmacro{\deltares}{#2-#1}%
  \pgfmathtruncatemacro{\valwhole}{int(#2)}%
  \pgfmathtruncatemacro{\valdec}{int(round((#2-\valwhole)*100))}%
  \ifnum\valdec<10\edef\valdecstr{0\valdec}\else\edef\valdecstr{\valdec}\fi%
  \ifdim\deltares pt>0pt%
    \pgfmathsetmacro{\deltaint}{int(min(100, abs(\deltares) * 1.8))}%
    \edef\deltacolor{deltagreen!\deltaint!white}%
    \begingroup\rmfamily\bfseries\small\expandafter\cellcolor\expandafter{\deltacolor}\valwhole.\valdecstr\endgroup%
  \else%
    \pgfmathsetmacro{\deltaint}{int(min(100, abs(\deltares) * 2))}%
    \edef\deltacolor{deltared!\deltaint!white}%
    \begingroup\rmfamily\bfseries\small\expandafter\cellcolor\expandafter{\deltacolor}\valwhole.\valdecstr\endgroup%
  \fi%
}

\newcommand{\chartdeltabgpctul}[2]{%
  \pgfmathsetmacro{\deltares}{#2-#1}%
  \pgfmathtruncatemacro{\valwhole}{int(#2)}%
  \pgfmathtruncatemacro{\valdec}{int(round((#2-\valwhole)*100))}%
  \ifnum\valdec<10\edef\valdecstr{0\valdec}\else\edef\valdecstr{\valdec}\fi%
  \ifdim\deltares pt>0pt%
    \pgfmathsetmacro{\deltaint}{int(min(100, abs(\deltares) * 1.8))}%
    \edef\deltacolor{deltagreen!\deltaint!white}%
    \expandafter\cellcolor\expandafter{\deltacolor}\small\rmfamily\myul{\valwhole.\valdecstr}%
  \else%
    \pgfmathsetmacro{\deltaint}{int(min(100, abs(\deltares) * 2))}%
    \edef\deltacolor{deltared!\deltaint!white}%
    \expandafter\cellcolor\expandafter{\deltacolor}\small\rmfamily\myul{\valwhole.\valdecstr}%
  \fi%
}

\newcommand{\chartdeltabgpctbful}[2]{%
  \pgfmathsetmacro{\deltares}{#2-#1}%
  \pgfmathtruncatemacro{\valwhole}{int(#2)}%
  \pgfmathtruncatemacro{\valdec}{int(round((#2-\valwhole)*100))}%
  \ifnum\valdec<10\edef\valdecstr{0\valdec}\else\edef\valdecstr{\valdec}\fi%
  \ifdim\deltares pt>0pt%
    \pgfmathsetmacro{\deltaint}{int(min(100, abs(\deltares) * 1.8))}%
    \edef\deltacolor{deltagreen!\deltaint!white}%
    \begingroup\rmfamily\bfseries\small\expandafter\cellcolor\expandafter{\deltacolor}\myul{\valwhole.\valdecstr}\endgroup%
  \else%
    \pgfmathsetmacro{\deltaint}{int(min(100, abs(\deltares) * 2))}%
    \edef\deltacolor{deltared!\deltaint!white}%
    \begingroup\rmfamily\bfseries\small\expandafter\cellcolor\expandafter{\deltacolor}\myul{\valwhole.\valdecstr}\endgroup%
  \fi%
}

\newcommand{\charttrainvaldeltabgpct}[2]{%
  \pgfmathsetmacro{\deltares}{#2-#1}%
  \pgfmathtruncatemacro{\valwhole}{int(#2)}%
  \pgfmathtruncatemacro{\valdec}{int(round((#2-\valwhole)*100))}%
  \ifnum\valdec<10\edef\valdecstr{0\valdec}\else\edef\valdecstr{\valdec}\fi%
  \ifdim\deltares pt>0pt%
    \pgfmathsetmacro{\deltaint}{int(min(100, abs(\deltares) * 5.2))}%
    \edef\deltacolor{deltagreen!\deltaint!white}%
    \expandafter\cellcolor\expandafter{\deltacolor}\small\rmfamily\valwhole.\valdecstr%
  \else%
    \pgfmathsetmacro{\deltaint}{int(min(100, abs(\deltares) * 2))}%
    \edef\deltacolor{deltared!\deltaint!white}%
    \expandafter\cellcolor\expandafter{\deltacolor}\small\rmfamily\valwhole.\valdecstr%
  \fi%
}

\newcommand{\charttrainvaldeltabgpctbf}[2]{%
  \pgfmathsetmacro{\deltares}{#2-#1}%
  \pgfmathtruncatemacro{\valwhole}{int(#2)}%
  \pgfmathtruncatemacro{\valdec}{int(round((#2-\valwhole)*100))}%
  \ifnum\valdec<10\edef\valdecstr{0\valdec}\else\edef\valdecstr{\valdec}\fi%
  \ifdim\deltares pt>0pt%
    \pgfmathsetmacro{\deltaint}{int(min(100, abs(\deltares) * 5.2))}%
    \edef\deltacolor{deltagreen!\deltaint!white}%
    \begingroup\rmfamily\bfseries\small\expandafter\cellcolor\expandafter{\deltacolor}\valwhole.\valdecstr\endgroup%
  \else%
    \pgfmathsetmacro{\deltaint}{int(min(100, abs(\deltares) * 2))}%
    \edef\deltacolor{deltared!\deltaint!white}%
    \begingroup\rmfamily\bfseries\small\expandafter\cellcolor\expandafter{\deltacolor}\valwhole.\valdecstr\endgroup%
  \fi%
}

\newcommand{\charttrainvaldeltabgpctul}[2]{%
  \pgfmathsetmacro{\deltares}{#2-#1}%
  \pgfmathtruncatemacro{\valwhole}{int(#2)}%
  \pgfmathtruncatemacro{\valdec}{int(round((#2-\valwhole)*100))}%
  \ifnum\valdec<10\edef\valdecstr{0\valdec}\else\edef\valdecstr{\valdec}\fi%
  \ifdim\deltares pt>0pt%
    \pgfmathsetmacro{\deltaint}{int(min(100, abs(\deltares) * 5.2))}%
    \edef\deltacolor{deltagreen!\deltaint!white}%
    \expandafter\cellcolor\expandafter{\deltacolor}\small\rmfamily\myul{\valwhole.\valdecstr}%
  \else%
    \pgfmathsetmacro{\deltaint}{int(min(100, abs(\deltares) * 2))}%
    \edef\deltacolor{deltared!\deltaint!white}%
    \expandafter\cellcolor\expandafter{\deltacolor}\small\rmfamily\myul{\valwhole.\valdecstr}%
  \fi%
}

\newcommand{\charttrainvaldeltabgpctbful}[2]{%
  \pgfmathsetmacro{\deltares}{#2-#1}%
  \pgfmathtruncatemacro{\valwhole}{int(#2)}%
  \pgfmathtruncatemacro{\valdec}{int(round((#2-\valwhole)*100))}%
  \ifnum\valdec<10\edef\valdecstr{0\valdec}\else\edef\valdecstr{\valdec}\fi%
  \ifdim\deltares pt>0pt%
    \pgfmathsetmacro{\deltaint}{int(min(100, abs(\deltares) * 5.2))}%
    \edef\deltacolor{deltagreen!\deltaint!white}%
    \begingroup\rmfamily\bfseries\small\expandafter\cellcolor\expandafter{\deltacolor}\myul{\valwhole.\valdecstr}\endgroup%
  \else%
    \pgfmathsetmacro{\deltaint}{int(min(100, abs(\deltares) * 2))}%
    \edef\deltacolor{deltared!\deltaint!white}%
    \begingroup\rmfamily\bfseries\small\expandafter\cellcolor\expandafter{\deltacolor}\myul{\valwhole.\valdecstr}\endgroup%
  \fi%
}

\definecolor{abstractbg}{HTML}{F2F7FB}
\definecolor{abstracturl}{HTML}{1C5EC2}
\renewenvironment{abstract}
  {\begin{tcolorbox}[colback=abstractbg,colframe=abstractbg,boxrule=0pt,arc=6pt,
                     left=14pt,right=14pt,top=10pt,bottom=10pt,boxsep=0pt,
                     width=\textwidth,grow to left by=24pt,grow to right by=24pt,
                     before skip=-0.1in,after skip=\medskipamount]
   \centerline{\large\sc Abstract}\vspace{0.6ex}}
  {\end{tcolorbox}\vskip 1ex}

\iclrfinalcopy 
\begin{document}

\maketitle

\begin{abstract}
Tackling complex real-world tasks can exceed the capabilities of a single large language model (LLM), motivating the use of multi-agent workflows that coordinate specialized agents to work together on these tasks.
Recent methods train LLMs to construct better workflows from execution outcomes, but they optimize only the workflow generator, while the other agents that build or execute each workflow remain fixed even though every outcome depends on all of them. However, extending training beyond the generator is challenging: the agents are coupled, and a workflow's outcome is a single sparse score that cannot tell which agent causes a failure. We propose \ours, which leverages the workflow as a harness to optimize workflows. By introducing a hierarchical, structure-aware reward paradigm, \ours enables one role to self-evolve and two or more roles to co-evolve, with no additional models, labels, or executions. Considering the limitation that workflows are commonly trained and evaluated on data that a single agent can already handle, we further propose \oursdata, an adaptive data hardening approach that converts existing datasets into workflow-level tasks with increased difficulty. Across document, slide, chart, code, math, and finance tasks, small open models trained with \ours achieve improved performance by up to $+7.41\%$, with co-evolving ($+5.03\%$) more roles gaining more than optimizing one of them alone ($+2.83\%$). Our project page: {\hypersetup{urlcolor=abstracturl}\url{https://xhguo7.github.io/FloWright/}}.
\end{abstract}

\section{Introduction}
\label{sec:intro}

Large language models (LLMs) are increasingly capable of leveraging tools and specialized skills to solve tasks~\citep{zhang2025nemotronresearchtooln1,qian2025toolrlrewardtoollearning,wang2025inducingprogrammaticskillsagentic,zhang2026landscapeagenticreinforcementlearning}. However, tackling complex real-world problems can go beyond the capability of a single LLM, as it demands gathering evidence across heterogeneous inputs, reasoning over long contexts and diverse sources, and accumulating intermediate results over many steps~\citep{yu2025visualdocumentunderstandingreasoning,zhu2025doclenstoolaugmentedmultiagent}. To address such tasks, LLMs construct multi-agent workflows that break complex tasks into subtasks and coordinate specialized agents~\citep{zhang2025aflow,hu2025adas,niu2025flowmodularizedagenticworkflow}.

Yet constructing an effective workflow for a complex task remains challenging, and many multi-agent frameworks still rely on substantial manual design~\citep{jin2026slideagent,li2025webweaver,agashe2025agents2}. To reduce this manual effort, recent works optimize LLMs to construct better workflows by training them against the outcomes of executing those workflows~\citep{nie2025weakforstrongtrainingweakmetaagent,nielsen2026learningorchestrateagents}.
However, these methods typically optimize a single workflow generator, while the agents that execute the workflow, and any other agents that help build it, remain fixed (\S\ref{subsec:related:generation}). Since the outcome of a workflow depends jointly on every agent that builds or executes it, \textit{training only the generator leaves other agents shaping every outcome without ever learning from it, significantly limiting how much the workflow as a whole can improve}.

Extending training beyond the generator introduces two additional challenges. First, agents in a workflow are interdependent: one agent's output can become another's input, and overall performance depends on their joint behavior Existing multi-agent training either trains each agent separately~\citep{guo2026anagentenhancingscientifictable,motwani2025malt} or co-trains agents with distinct roles only within a fixed pairing or a hand-designed system~\citep{park2025maporl,zhao2026strongermas,huang2026rzero,wan2025rema} (\S\ref{subsec:related:multirole}).
%
\textit{Training agents separately thus ignores how each agent's behavior shapes what the others learn, while co-training remains confined to a single role or a predefined agent structure rather than the workflows constructed for each task}.
Second, the outcome of a workflow is usually a single score for the whole workflow. It is sparse whenever the workflow falls short, and cannot tell which agent causes a failure. Methods that do attribute failures to specific agents or steps, however, rely heavily on learned critics or reward models, LLM judges, human annotation, or repeated executions~\citep{liu2025agenticreinforcementlearningimplicit,zhang2025agentracer,ma2026dover} (\S\ref{subsec:related:credit}).
\textit{Such attribution adds costly models, labels, or executions beyond the workflow run itself, and it can still misattribute where a failure originates}.

Beyond how workflows are built and trained, the data used to optimize and evaluate them pose a further challenge. Multi-agent workflows are commonly trained and evaluated on datasets such as single-question math, function-level coding, and short question answering, which a single agent can already handle~\citep{zhang2026rethinkingevaluationefficiencymethods} (\S\ref{subsec:related:benchmarks}). On such data, re-evaluations find that the gains of multi-agent systems over a strong single agent are often small or absent~\citep{cemri2025multiagentllmsystemsfail,choi2025debatevoteyieldsbetter,zhu2026demystifyingmultiagentdebaterole} (\S\ref{appendix:sec:preliminary}). \textit{These datasets can therefore neither reveal nor optimize what a workflow adds beyond a single agent}.

Motivated by these limitations, we propose \ours (\S\ref{sec:method}), which leverages the workflow as a harness (\S\ref{sec:prelim}) to optimize any role (\S\ref{subsec:method:objective}). Besides letting each role self-evolve, \ours co-evolves multiple roles inside the harness (\S\ref{subsec:method:evolve}), and stages the grade into a hierarchical, structure-aware reward (\S\ref{subsec:method:reward}) with role-level credit localization (\S\ref{subsec:method:credit}). Beyond workflow construction and optimization, \ours introduces a data adaptation algorithm (Alg.~\ref{alg:data_adaptation}) that hardens existing datasets for workflow-level training and evaluation (\S\ref{appendix:sec:data_adaptation}).
Our contributions are as follows:
\begin{oursenumerate}
\item \textbf{Shared harness for every role, at train and test time.} We define a shared workflow harness whose signal serves three purposes (\S\ref{subsec:prelim:harness}): the evaluation metric (\S\ref{appendix:subsec:data_adaptation:eval_metrics}), the optimization signal for training any role (Fig.~\ref{fig:flowright_main}), and the meta distillation objective at test time (\S\ref{subsec:method:objective}).
\item \textbf{Structure-aware credit at no extra cost.} We design a reward hierarchy that turns the sparse signal of a workflow into dense, role-level feedback (\S\ref{subsec:method:reward}), with credit read from the execution trace that the harness already records, requiring no additional models, labels, or executions (\S\ref{subsec:method:credit}).
\item \textbf{Workflow-level data hardening.} We propose \oursdata, a data adaptation approach whose three hardening strategies convert datasets a single agent can already handle into workflow-level tasks with increased difficulty levels, supporting both training and evaluation (\S\ref{appendix:sec:data_adaptation}).
\item \textbf{Four evolution modes that generalize.} \ours enables four evolution modes (\S\ref{subsec:method:evolve}): (a) single-role self-evolution, (b) agent-skill co-evolution, (c) upstream-downstream co-evolution, and (d) multi-agent co-evolution. Co-evolving more roles
gains more than optimizing one alone, achieving $+5.03\%$ overall against $+2.83\%$ for a single role (\S\ref{subsec:results:main}). In addition, gains are generalizable across roles, backbones, RL algorithms, workflow topologies, and pools
(\S\ref{subsec:results:ablation}).
\end{oursenumerate}

\section{Preliminaries}
\label{sec:prelim}


\subsection{Workflow as Harness}
\label{subsec:prelim:harness}

A complex task $x \sim \mathcal{X}$ is solved by a \textit{workflow}, a directed graph $G = (V, E)$ whose nodes $V$ are agents, tools, or skills, and whose edges $E$ carry the data and control flow among them. A \textit{harness} $\mathcal{H}$ grounds a workflow $G$ into verifiable measurement through execution and grading: it runs $G$ on $x$, returns the answer $y$, and scores it as $h(G, x) \in [0, 1]$. A \textit{Generator} $\pi_g$ produces $G$ for $x$:
\begin{equation}
    x \;\xrightarrow{\;\pi_g\;}\; G \;\xrightarrow{\;\mathcal{H}\;}\; \big(y,\; h(G, x)\big), \quad h(G, x) \in [0, 1]
    \label{eq:pipeline}
\end{equation}
where $h$ serves three purposes: the evaluation metric, the training signal, and the test-time objective. A \textit{role} $\rho \in \mathcal{R}$ (Tab.~\ref{tab:notation}) is any component of the upstream-downstream framework (Fig.~\ref{fig:upstream_downstream_structure}) whose policy $\pi_\rho$, with parameters $\theta_\rho$, is optimized against that signal.

\subsection{Upstream Builds, Downstream Runs}
\label{subsec:prelim:parts}

\begin{figure}[!t]

    \small
    \centering
    \includegraphics[width=1.0\textwidth]{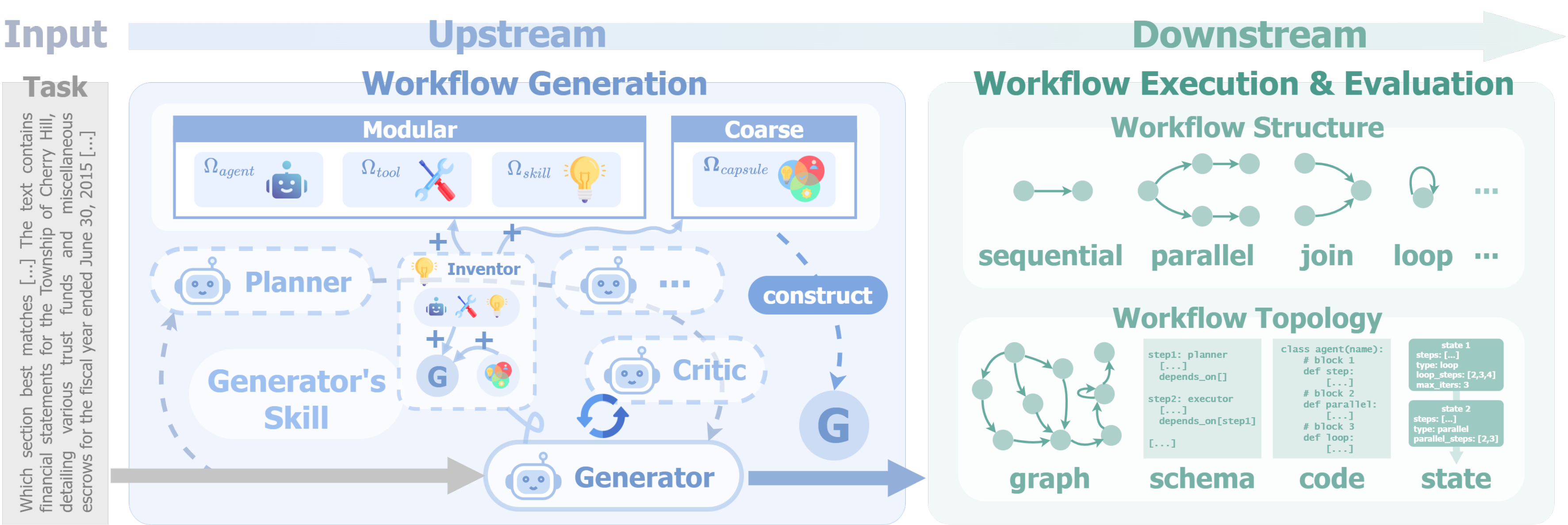}
    
    \vspace{-6pt}
    
    \caption{\textbf{Upstream-Downstream Framework.} Upstream, workflow generation can take on different structures, from \textit{Generator} with its own \textit{Inventor} skill to \textit{Generator} working with other agents, drawing on pools of different granularity. Downstream, workflows take different structures and topologies to better solve the task.}
    \label{fig:upstream_downstream_structure}

    \vspace{-3pt}

\end{figure}

\ours factors into an \textit{upstream} and a \textit{downstream}. Upstream, the \textit{Generator} $\pi_g$ produces $G$. As shown in Fig.~\ref{fig:upstream_downstream_structure}, upstream workflow generation can take on different structures to construct a better $G$. Downstream, the harness $\mathcal{H}$ executes and grades $G$ to return $\big(y,\, h(G, x)\big)$. The workflow is the interface between them: \textit{what the upstream constructs is what the downstream executes} (Eq.~\ref{eq:pipeline}).

\subsection{What the Nodes Are Made of}
\label{subsec:prelim:pool}

A workflow is built from a shared \textit{pool} $\Omega$ of reusable agents, tools, and skills. Each node of $G$ is an agent, a tool, or a skill from $\Omega$, so $\Omega$ delimits the space of workflows the \textit{Generator} can produce.

\section{\ours: Learning to Wright Workflows}
\label{sec:method}

For a task $x_i \in \mathcal{X}$, upstream produces a workflow $G \in \mathcal{G}$ (\S\ref{sec:prelim}, \ref{subsec:method:wire}). \ours (Fig.~\ref{fig:flowright_main}) grounds it into a measurable \textit{optimization signal} $h(G, x_i)$ through a principled harness with both local crediting (\S\ref{subsec:method:credit}) and a global hierarchy (\S\ref{subsec:method:reward}). By learning to optimize this signal, \ours enables LLMs to solve complex tasks by building better workflows, through self-evolution and co-evolution (\S\ref{subsec:method:evolve}) at both train and test time (\S\ref{subsec:method:objective}).

\subsection{Wire the Workflow: Topology \& Pool Dynamics}
\label{subsec:method:wire}

Upstream construction turns on three choices: the topology of the workflow $G$, and the granularity and dynamics of the pool $\Omega$ it draws from.

\textbf{Topology is a non-linear control-flow \textit{graph}.}
Beyond passing progress forward in a chain, $\pi_g$ runs nodes in parallel when subtasks are independent, forks alternative branches that are later reconciled, and repeats a step until a stopping condition is met. A single workflow can thus express sequential, parallel, branching, and iterative structure as the task demands.

\textbf{Pool is presented to the upstream at a chosen \textit{granularity} (Fig.~\ref{fig:upstream_downstream_structure}).}
A fine, modular granularity keeps agents, tools, and skills as separate units to enable flexible selection and composition: $\Omega = \Omega_{\textit{agent}} \cup \Omega_{\textit{tool}} \cup \Omega_{\textit{skill}}$.
A coarse granularity bundles each agent with necessary tools and skills into a self-contained capsule, leading to $\Omega = \Omega_{\textit{capsule}}$.

\textbf{Pool is \textit{dynamic}.}
When a task calls for a capability that $\Omega$ does not contain, the pool is expanded with the missing component, a new tool, skill, or agent. The upstream thus grows its own components during construction in addition to drawing directly from the pool (Fig.~\ref{fig:upstream_downstream_structure}). Holding the pool static disables this growth, a setting we ablate in our experiments (\S\ref{subsec:results:ablation}).

\subsection{Credit the Structure: Localizing Credits to Roles}
\label{subsec:method:credit}

For the $j$-th workflow $G_j = (V_j, E_j)$ executed on $x_i$, the optimization signal $h(G_j, x_i)$ is a single scalar, but the action it measures spans an entire graph. By \textit{harnessing the harness}, \textit{i.e.}, leveraging the execution trace $\tau_j$ produced by $\mathcal{H}$, we localize it to the roles responsible. Credit localization therefore requires no additional execution and no new signal.

\begin{figure}[!t]

    \small
    \centering
    \includegraphics[width=1.0\textwidth]{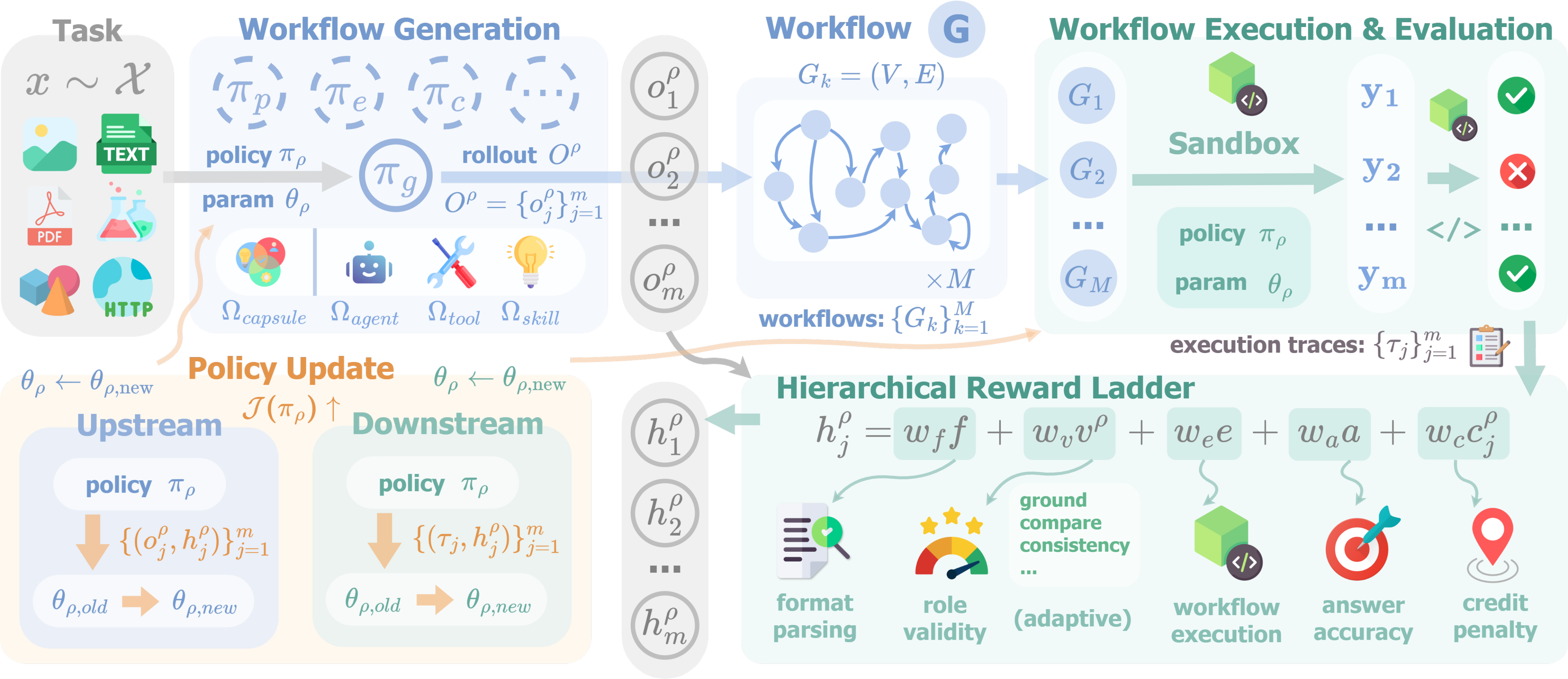}
    
    \vspace{-6pt}
    
    \caption{\textbf{\ours for reinforcement learning via self-evolving harness.} $\pi_\rho$ is the policy of a role $\rho \in \mathcal{R}$. For the $j$-th of $m$ rollouts of a role being optimized, $G_j \in \mathcal{G}$ is the workflow that rollout is graded on, and the $m$ rollouts span $M$ distinct workflows (Tab.~\ref{tab:notation}). Workflow generation and execution form a coupled on-policy learning loop: upstream proposes workflows, downstream executes them, and the verifiable harness turns execution outcomes into learning signals for the policies being optimized. Policy updates produce the next generation of workflows and executions, enabling self-evolution by harnessing the system's own experience without additional supervision.}
    \label{fig:flowright_main}

    \vspace{0pt}

\end{figure}

\textbf{Credit charges each role for the failures of its own nodes.}
A role $\rho$ authors a subset $V_j^\rho \subseteq V_j$ of the workflow's nodes (Fig.~\ref{fig:flowright_main}), \textit{e.g.}, \textit{Generator} authors every node it designs and wires, \textit{Inventor} authors only the nodes that use a component it creates, and \textit{Downstream} only the agent nodes it runs. From $\tau_j$ we recover $F_j^\rho \subseteq V_j^\rho$, the role's own nodes that root-cause a failure, and charge the role in proportion to the fraction of its footprint at fault:
\begin{equation}
c_j^\rho = -\frac{|F_j^\rho|}{|V_j^\rho|} \in [-1, 0]
\label{eq:credit}
\end{equation}
If $F_j^\rho=\varnothing$, \textit{e.g.}, when no node fails, then $c_j^\rho = 0$; otherwise the term is negative, steering the policy away from failures (\S\ref{appendix:subsec:method_extension:credit}).

\subsection{Harness the Workflow: A Hierarchical Reward}
\label{subsec:method:reward}

The harness does not reduce an execution to a single pass-or-fail outcome. Instead, it stages the optimization signal into a principled hierarchy of auto-verified terms, so the policy sees a meaningful gradient even when the workflow falls short.
As such, we propose our hierarchical reward as a generic paradigm: one \textit{shared harness} from which every optimizable role adaptively instantiates its own reward.

\textbf{Optimization signal is a staged ladder.}
Reward terms accumulate from the role's own output to the task outcome: whether the role's output $o_j^\rho$ \textit{parses} ($f$), whether the role's contribution is \textit{valid} ($v^\rho$), whether the workflow \textit{executes} to a final answer ($e$), and how correct that \textit{answer} is ($a$). The local credit of \S\ref{subsec:method:credit} enters as a fifth, structure-aware term, extending the ladder across local and global.
As such, the signal for role $\rho$ is:
\begin{equation}
h_j^\rho = h^\rho(G_j, x_i) = w_f\, f + w_v\, v^\rho + w_e\, e + w_a\, a + w_c\, c_j^\rho
\label{eq:hierarchical_reward}
\end{equation}
where $\{w_f, w_v, w_e, w_a, w_c\}$ are weighting coefficients configured per role (\S\ref{appendix:subsec:implementation:config}).
Our hierarchical ladder thus composes one optimization signal across two axes: \textit{from the role's own output to the task outcome}, the staged four terms measure how far a workflow reaches; \textit{from local to global}, the credit term (\S\ref{subsec:method:credit}) localizes failures to roles at fault. \ours aims to maximize this signal at train and test time (\S\ref{subsec:method:evolve}, \ref{subsec:method:objective}).

\subsection{Harness for Evolution: One Role or Mutual Curriculum}
\label{subsec:method:evolve}

Our reward paradigm serves every role from one shared harness, enabling one role to self-evolve or several to co-evolve.

\textbf{Any single role self-evolves inside the harness.}
To train a role, we optimize its role-specific reward $h^\rho$ inside the harness as it runs, so the role evolves in the same flow that will use it (\S\ref{sec:exps}a). The same procedure fits any role, since all roles share one harness.

\textbf{Several roles co-evolve inside the harness through a mutual curriculum.}
When several roles train on one substrate inside one harness, each role's behavior shapes what the others encounter. Improving one role therefore reshapes the experience the rest learn from. The co-evolving roles can thus share one policy or each keep their own, and in either case each role learns against the latest evolved policies of the others. For example, co-evolution can pair an agent with its own skill (\S\ref{sec:exps}b), couple an upstream agent that builds the workflow with a downstream agent that executes it (\S\ref{sec:exps}c), or co-train any two or more agents (\S\ref{sec:exps}d). In each case, the coupling is a \textit{mutual curriculum}: each role shapes the training distribution the others learn from. The structure-aware credit of \S\ref{subsec:method:credit} charges each role only for the failures of its own nodes in the shared execution, which separates the roles' credit wherever their footprints differ.

\subsection{Maximize the Harness: Train-Time RL, Test-Time Distillation}
\label{subsec:method:objective}

\ours optimizes the signal in two stages: train-time reinforcement and test-time distillation.

\textbf{Train time: maximizing the signal.}
We optimize a role's policy $\pi_\rho$ to maximize its expected signal over tasks:
\begin{equation}
\mathcal{J}(\pi_\rho) = \mathbb{E}_{x_i \sim \mathcal{X},\; \{\pi_{\rho'}\}_{\rho' \in \mathcal{R}}}\big[\, h^\rho(G_j, x_i) \,\big]
\label{eq:objective}
\end{equation}
where $\mathcal{J}$ optimizes $\pi_\rho$ while the expectation runs over the policies $\pi_{\rho'}$ of all roles $\rho' \in \mathcal{R}$ in the flow.
As \ours contributes a training paradigm rather than an optimizer, $\mathcal{J}$ admits any policy-gradient method: we anchor on the proximal policy optimization~\citep{schulman2017ppo} clipped surrogate and leave the advantage estimator and policy loss interchangeable, so recent variants such as GRPO~\citep{shao2024grpo}, DAPO~\citep{yu2025dapo}, and CISPO~\citep{minimax2025cispo} all instantiate it, and we ablate the choice in our experiments (\S\ref{subsec:results:ablation}).

\textbf{Test time: meta distillation without gradients.}
The same shared harness also optimizes workflows at test time, where no weights change. In place of the policy parameters, \ours optimizes a reusable prior $P$ that optimize generation:
\begin{equation}
P^\star = \arg\max_{P}\; \mathbb{E}_{x_i \sim \mathcal{X}}\big[\, h(G, x_i \mid P) \,\big]
\label{eq:distill}
\end{equation}
This prior is distilled \textit{internally} from the system's own experiences, or \textit{externally} from a stronger teacher, or both. We additionally distill successful experiences into few-shot cases, from which roles evolve by \textit{meta learning}. We define this \textit{meta distillation}: the harness turns what its own successful experiences teach into the prior that later tasks reuse. As such, \ours keeps improving on new tasks, driven by the very signal it optimizes during training.

\section{Experiments}
\label{sec:exps}

\textbf{Setup.}
We evaluate \ours in both objectives it optimizes (\S\ref{subsec:method:objective}): reinforcement learning at train time, and meta distillation at test time. At train time, we cover four evolution modes (\S\ref{subsec:method:evolve}): (a) single-role self-evolution, (b) agent-skill co-evolution, (c) upstream-downstream co-evolution, and (d) multi-agent co-evolution. We compare them against two single-agent baselines, with or without the tool and skill pools (Fig.~\ref{fig:upstream_downstream_structure}). Each baseline runs with the same compute budget as our \ours counterpart it is compared against. All evaluations are on the held-out test sets, non-overlapped with training data (\S\ref{appendix:subsec:data_adaptation:stats}). We train \texttt{Qwen3.5-4B} and \texttt{Qwen3.5-9B}~\citep{qwen2026qwen35omnitechnicalreport} on the \textit{paired} strategy at $\ell=3$, and evaluate two open-source models, \texttt{Qwen3.5-4B} and \texttt{Qwen3.5-9B}, together with two close-source models, \texttt{GPT-5-mini}~\citep{openai2025gpt5mini} and \texttt{GPT-5.4-mini}~\citep{openai2026gpt54mini}, on held-out test sets, using \texttt{GPT-5.4}~\citep{openai2026gpt54} as the distillation teacher (\S\ref{subsec:method:objective}) and \texttt{GPT-5.6-luna}~\citep{openai2026gpt56luna} as LLM judge (\S\ref{appendix:subsec:data_adaptation:eval_metrics}). As a role can be powered by a different backbone than the rest, we additionally evaluate settings that pair a smaller upstream backbone with a larger downstream one as well as the reverse. We summarize our implementation details in \S\ref{appendix:sec:implementation}.

\textbf{Data.}
We evaluate on the hardened tasks of \oursdata (\S\ref{appendix:sec:data_adaptation}), which converts $12$ datasets over $7$ domains into workflow-level tasks under three hardening strategies (\S\ref{appendix:subsec:data_adaptation:strategies}), at hardening levels $\ell \in \{3, 5\}$. Each dataset is split from the source before hardening to preclude train-test overlap. Tab.~\ref{tab:data_stats} summarizes \oursdata data statistics. We refer to each $<\textit{dataset}, \textit{hardening strategy}, \textit{hardening level}>$ triple as an evaluation \textit{arm}, yielding $44$ arms in total (Tab.~\ref{tab:exp_main}, \S\ref{appendix:subsec:data_adaptation:stats}). Each arm falls into one of three evaluation \textit{regimes} by its distance from the arms \ours trains on: \textit{in-distribution}, \textit{out-of-distribution}, and \textit{out-of-domain} (Tab.~\ref{tab:exp_main}).

\textbf{Evaluation Metrics.}
A hardened task is scored as the mean over its sub-questions, and a dataset's accuracy is the mean over its test tasks (\S\ref{appendix:subsec:data_adaptation:eval_metrics}). Every sub-question keeps the metric of its own answer type (Tab.~\ref{tab:metrics}), so bundling changes how hard a task is without changing how its answers are graded.


\section{Results}
\label{sec:results}

\begin{table*}[t]

\centering
\caption{\textbf{Evaluation on Four Evolution Modes.} Mean accuracy (\%) over the hardening strategies and levels each dataset supports (\S\ref{appendix:subsec:data_adaptation:strategies}). \ours trains on the \textit{paired} strategy at $\ell{=}3$ of four datasets, which defines three regimes (\S\ref{appendix:subsec:data_adaptation:stats}): \colorbox{tea_bg}{\textcolor{tea_text}{\textbf{in-distribution}}} is those four training arms; \colorbox{blue_bg}{\textcolor{blue_text}{\textbf{out-of-distribution}}} is the same four datasets under a strategy, a level, or both that training never sees (13 arms); and \colorbox{rose_bg}{\textcolor{rose_text}{\textbf{out-of-domain}}} is eight datasets \ours never trains on (27 arms). In-distribution is $\ell{=}3$ by construction, since $\ell{=}3$ is the training level and $\ell{=}5$ is held out. The \colorbox{tea_delta!22}{green shading} is proportional to the gain over untrained backbones. \texttt{4B} and \texttt{9B} abbreviate \texttt{Qwen3.5-4B} and \texttt{Qwen3.5-9B}. (a)-(d) denote the four evolution modes defined in \S\ref{sec:exps}.}
\label{tab:exp_main}

\vspace{3pt}

\small

\renewcommand{\arraystretch}{1.2}
\newcommand{\blockgap}{\noalign{\vskip 2pt}}
\setlength{\tabcolsep}{2.5pt}

\begin{tabularx}{\linewidth}{@{}>{\bfseries}l*{8}{>{\centering\arraybackslash}X}@{}}
\toprule
\multicolumn{1}{@{}c}{\multirow{2}{*}{\textbf{Method}}} & \multicolumn{1}{c}{\textbf{In-Dist.}} & \multicolumn{3}{c}{\textbf{Out-of-Distribution}} & \multicolumn{3}{c}{\textbf{Out-of-Domain}} & \multicolumn{1}{c@{}}{\multirow{2}{*}{\textbf{Overall}}} \\
\cmidrule(lr){2-2}\cmidrule(lr){3-5}\cmidrule(lr){6-8}
 & \cellcolor{tea_bg}\textcolor{tea_text}{$\bm{\ell{=}3}$} & \cellcolor{blue_bg}\textcolor{blue_text}{$\bm{\ell{=}3}$} & \cellcolor{blue_bg}\textcolor{blue_text}{$\bm{\ell{=}5}$} & \cellcolor{blue_bg}\textcolor{blue_text}{\textbf{Mean}} & \cellcolor{rose_bg}\textcolor{rose_text}{$\bm{\ell{=}3}$} & \cellcolor{rose_bg}\textcolor{rose_text}{$\bm{\ell{=}5}$} & \cellcolor{rose_bg}\textcolor{rose_text}{\textbf{Mean}} & \\
\midrule

\rowcolor{gray_bg}
\multicolumn{9}{@{}c@{}}{\textcolor{gray_text}{\textbf{Baseline: Single Agent (\textit{without tool or skill})}}} \\

\blockgap

Qwen3.5-4B                             & 10.87 & 11.39 & 6.12 & 8.15 & 7.89 & 5.12 & 6.56 & 7.42 \\
Qwen3.5-9B                             & 16.17 & 14.45 & 9.35 & 11.31 & 13.55 & 9.53 & 11.61 & 11.94 \\

\blockgap

\rowcolor{gray_bg}
\multicolumn{9}{@{}c@{}}{\textcolor{gray_text}{\textbf{Baseline: Single Agent (\textit{with tools and skills})}}} \\

\blockgap

Qwen3.5-4B                             & 21.13 & 20.75 & 15.12 & 17.28 & 25.02 & 19.35 & 22.29 & 20.70 \\
Qwen3.5-9B                             & 28.53 & 31.14 & 23.87 & 26.67 & 35.79 & 30.08 & 33.04 & 30.75 \\

\blockgap

\rowcolor{blue_bg}
\multicolumn{9}{@{}c@{}}{\textcolor{blue_text}{\textbf{Baseline: \textsc{FloWright} (\textit{untrained})}}} \\
\blockgap

Qwen3.5-4B                             & 31.91 & 34.91 & 25.70 & 29.24 & 36.17 & 30.58 & 33.48 & 32.09 \\
Qwen3.5-9B                             & 50.14 & 44.47 & 40.48 & 42.02 & 52.44 & 46.84 & 49.74 & 47.50 \\

\blockgap

\rowcolor{tea_bg}
\multicolumn{9}{@{}c@{}}{\textcolor{tea_text}{\textbf{\textsc{FloWright}: Train-Time Optimization}}} \\
\blockgap
\;\; + (a) 4B                  & \dcell{17}{35.16} & \dcell{15}{37.78} & \dcell{18}{29.15} & \dcell{17}{32.47} & \dcell{11}{38.38} & \dcell{15}{33.58} & \dcell{13}{36.07} & \dcell{15}{34.92} \\
\;\; + (a) 9B                  & \dcell{22}{\textbf{54.51}} & \dcell{20}{\textbf{48.28}} & \dcell{16}{\textbf{43.66}} & \dcell{18}{\textbf{45.44}} & \dcell{23}{\textbf{56.99}} & \dcell{18}{\textbf{50.36}} & \dcell{21}{\textbf{53.80}} & \dcell{20}{\textbf{51.39}} \\
\;\; + (b) 4B                  & \dcell{38}{39.32} & \dcell{24}{39.50} & \dcell{27}{30.94} & \dcell{26}{34.23} & \dcell{16}{39.27} & \dcell{16}{33.74} & \dcell{16}{36.61} & \dcell{21}{36.15} \\
\;\; + (c) 4B                  & \dcell{22}{36.23} & \dcell{20}{38.75} & \dcell{21}{29.88} & \dcell{21}{33.29} & \dcell{14}{38.89} & \dcell{11}{32.65} & \dcell{12}{35.89} & \dcell{16}{35.15} \\
\;\; + (d) 4B                  & \dcell{38}{39.23} & \dcell{30}{40.71} & \dcell{32}{31.91} & \dcell{31}{35.29} & \dcell{22}{40.48} & \dcell{21}{34.67} & \dcell{22}{37.68} & \dcell{26}{37.12} \\
\bottomrule
\end{tabularx}

\end{table*}

\subsection{Evolve the Roles: Optimize Any Role by Harnessing the Workflow}
\label{subsec:results:main}

\textbf{\ours improves performance in every regime at train time.} Comparing with single-agent baselines, untrained \ours showcases improved performance ($\geq+11.39\%$ overall). This reveals that models with smaller sizes are able to build and run workflows with gains better than its size acting alone. At matched model size, every evolution mode exceeds its untrained counterpart in every regime and overall by up to $+7.41\%$ (Tab.~\ref{tab:exp_main}), and every arm improves under at least one mode. Moreover, the gain also increases when more roles co-evolve. Multi-agent co-evolution gains $+7.32\%$ in-distribution, $+6.05\%$ out-of-distribution, and $+4.20\%$ out-of-domain, staying positive across all datasets, strategies, and levels (Fig.~\ref{fig:exp_heterogeneity}). Among the four \texttt{Qwen3.5-4B} modes, multi-agent co-evolution leads $7/8$ columns of Tab.~\ref{tab:exp_main} and gains most overall at $+5.03\%$, ahead of agent-skill co-evolution at $+4.06\%$, upstream-downstream co-evolution at $+3.06\%$ and single-role self-evolution at $+2.83\%$. These improvements indicate that co-evolving more roles yields more than optimizing one of them alone. Harnessing the workflow therefore carries what \ours learns on four datasets at one hardening strategy and one level to the strategies, levels, and domains held out from it.

\begin{figure}[t]
    
    \centering
    \includegraphics[width=1.0\textwidth]{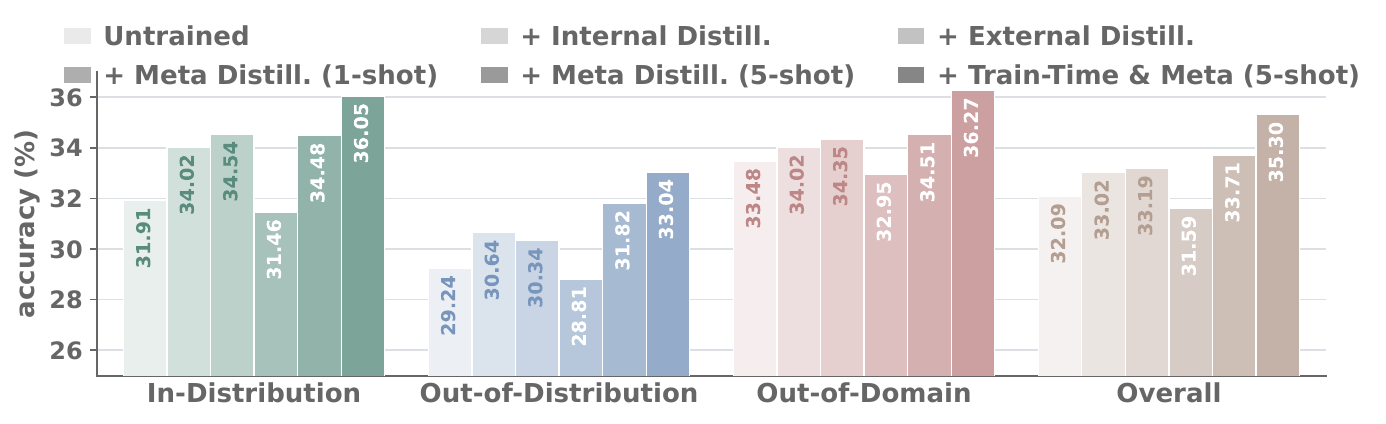}
    
    \vspace{-10pt}

    \caption{\textbf{Test-Time Optimization with Reusable Prior.} The leftmost bar of each group is the untrained \texttt{Qwen3.5-4B} workflow baseline, and every other bar generates workflows by distilling a reusable prior (\S\ref{subsec:method:objective}). \textit{Internal} and \textit{external} distillation draw that prior from own experiences and from a stronger teacher (GPT-5.4) respectively. \textit{Meta} distillation additionally distills successful experiences into few-shot cases, from which roles self-evolve via meta learning (1-shot and 5-shot). The rightmost bar composes train-time evolution with the 5-shot prior.}
    \label{fig:exp_test_time}
    
    \vspace{20pt}
    
\end{figure}

\textbf{\ours also improves performance at test time.} In place of policy parameters, \ours additionally distills a reusable prior at test time (\S\ref{subsec:method:objective}) for improved adaptability on new tasks. Distilling this prior from its own experiences gains $+0.93\%$ overall, and from a stronger teacher (GPT-5.4) $+1.10\%$ (Fig.~\ref{fig:exp_test_time}). Meta distillation gives the largest test-time gain at five shots with up to $+2.58\%$, in contrast to one shot staying below the untrained baseline. \textit{A prior helps only when it carries enough evidence to generalize beyond the single case it comes from.} Test-time and train-time optimization then compound, reaching $+3.21\%$ overall, $+4.14\%$ in-distribution, $+3.80\%$ out-of-distribution, and $+2.79\%$ out-of-domain, exceeding the five-shot prior alone by  $+1.59\%$ and leading other settings across regimes. The same harness (\S\ref{subsec:method:objective}) therefore optimizes workflows whether or not the policy weights update.

\textbf{\ours generalizes across roles and backbones at both train and test time.} At train time, optimizing any single role improves the performance, and co-evolving more roles improves it further (Fig.~\ref{fig:exp_train_roles}). Also, an optimized \texttt{4B} role is able to transfer its gain when other roles run on \texttt{Qwen3.5-9B}, even raising accuracy above the all-\texttt{9B} baseline (Fig.~\ref{fig:exp_mix_model}). At test time, the meta-distilled prior demonstrates persistent gain in every regime on all four open-source and close-source model backbones (Fig.~\ref{fig:exp_gpt_models}). \textit{One harness therefore optimizes any role and backbone a workflow is built and executed upon}. We extend our discussion in \S\ref{appendix:subsec:results:generalization}.

\vspace{12pt}

\subsection{Ablate the Design: Algorithm, Reward, Pool, and Topology}
\label{subsec:results:ablation}

\vspace{6pt}

\begin{wrapfigure}[12]{r}{0.5\textwidth}
    \vspace{-16pt}
    \centering
    \includegraphics[width=0.48\textwidth]{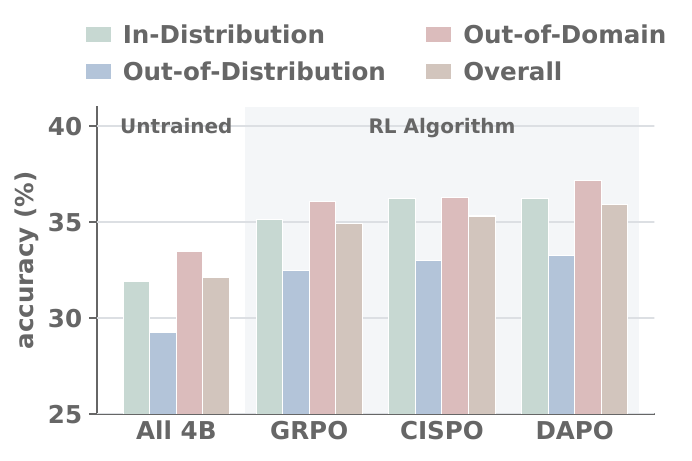}

    \vspace{-15pt}

    \caption{\textbf{Algorithm Generalizability.} Each algorithm optimizes \textit{Generator} on \texttt{Qwen3.5-4B}.}
    \label{fig:exp_rl_algorithm}

    \vspace{0pt}
\end{wrapfigure}

\textbf{\ours is generalizable to different RL algorithms.}
Optimizing \textit{Generator} consistently outperforms the untrained baseline with different RL algorithms (Fig.~\ref{fig:exp_rl_algorithm}): $+2.83\%$ overall for GRPO, $+3.23\%$ for CISPO, and $+3.84\%$ for DAPO. DAPO leads in every regime, and its margin widens in out-of-domain evaluation, where it gains $+3.69\%$ against $+2.82\%$ for CISPO and $+2.59\%$ for GRPO. On the other hand, for in-distribution, it stays within $0.01\%$ of CISPO. \textit{The harness provides a learning signal generalizable to different RL algorithms.}

\clearpage
\begin{wrapfigure}[11]{r}{0.5\textwidth}
    \vspace{-12pt}
    \centering
    \includegraphics[width=0.48\textwidth]{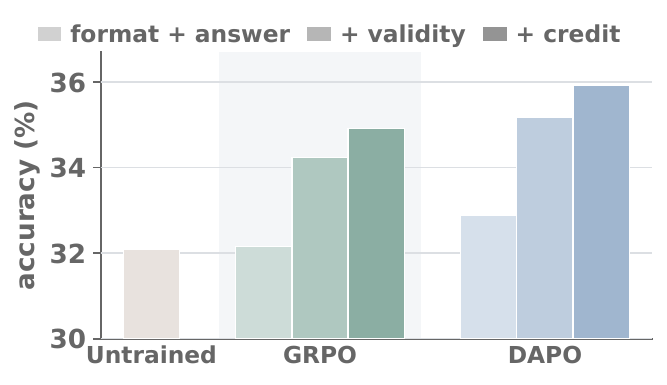}

    \vspace{-15pt}

    \caption{\textbf{Reward Ladder.} Each layer adds one term (\S\ref{subsec:method:reward}): format and answer alone, plus validity (\S\ref{appendix:subsec:method_extension:validity}), plus the structure-aware credit (\S\ref{subsec:method:credit}).}
    \label{fig:exp_reward_ladder}

    \vspace{0pt}
\end{wrapfigure}

\textbf{Every layer of the reward ladder contributes.} As shown in Fig.~\ref{fig:exp_reward_ladder}, grading format and answer alone stays close to the untrained baseline, with $+0.07\%$ under GRPO and $+0.80\%$ under DAPO. Adding validity reward contributes the largest single increment at $+2.07\%$ under GRPO and $+2.29\%$ under DAPO, respectively. The structure-aware credit of \S\ref{subsec:method:credit} adds a further $+0.69\%$ and $+0.75\%$, reaching $+2.83\%$ and $+3.84\%$ overall, respectively. \textit{Each layer of the reward ladder contributes, with validity contributing the most.}

\begin{wrapfigure}[12]{r}{0.5\textwidth}
    \vspace{-5pt}
    \centering
    \includegraphics[width=0.46\textwidth]{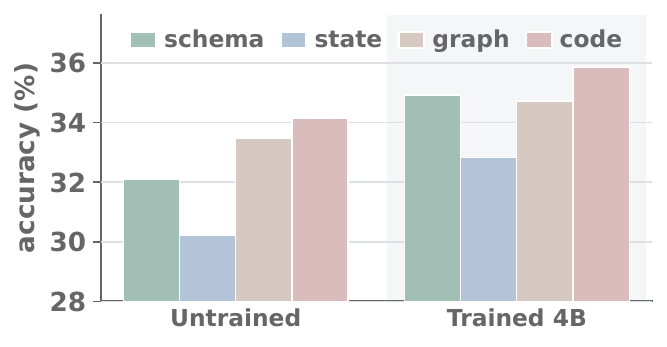}

    \vspace{-10pt}

    \caption{\textbf{Workflow Topology.} \textit{Generator} writes workflows in schema, state machine, graph, or code (\S\ref{appendix:subsec:method_extension:topologies}). Both use the \texttt{Qwen3.5-4B} backbone.}
    \label{fig:exp_topology}

\end{wrapfigure}

\textbf{\ours is generalizable to different workflow topologies.} As shown in Fig.~\ref{fig:exp_topology}, using \textit{code} to represent workflows reaches $34.12\%$ untrained and $35.86\%$ with an optimized \textit{Generator}, ahead of the other three representations in both. Optimization improves every topology, by $+2.83\%$ for \textit{schema}, $+2.62\%$ for \textit{state machine}, $+1.74\%$ for \textit{code}, and $+1.25\%$ for \textit{graph}, narrowing the spread across the four from $3.91\%$ to $3.03\%$. \textit{Optimization improves across workflow representations, so the gains hold whichever topology the \textit{Generator} writes in.}

\begin{wrapfigure}[15]{r}{0.5\textwidth}
    \vspace{-10pt}
    \centering
    \includegraphics[width=0.48\textwidth]{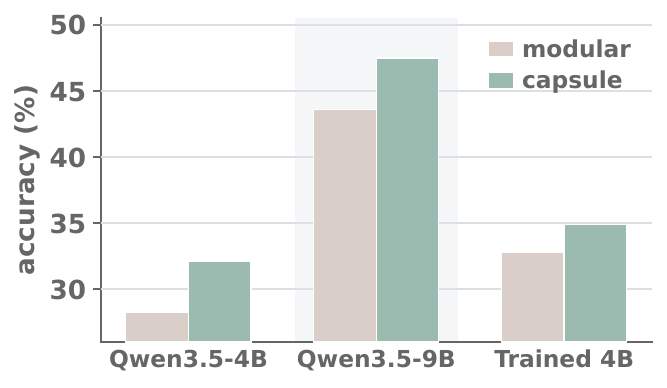}

    \vspace{-10pt}

    \caption{\textbf{Pool Granularity.} Backbones are compared at two pool granularities (\S\ref{subsec:method:wire}), and \textit{Trained 4B} carries RL-trained \textit{Generator}.}
    \label{fig:exp_pool_granularity}

    \vspace{-10pt}
\end{wrapfigure}

\textbf{Capsules lead at every backbone, and optimization narrows the gap.}
As shown in Fig.~\ref{fig:exp_pool_granularity}, bundling each agent with the tools and skills it needs gains $+3.83\%$ on \texttt{Qwen3.5-4B} and $+3.87\%$ on \texttt{Qwen3.5-9B} over a modular pool (\S\ref{subsec:method:wire}). Optimizing \textit{Generator} improves the modular pool by $+4.50\%$ against $+2.83\%$ for capsules, narrowing the gap between the two to $+2.16\%$. \textit{Composing a workflow from separate units demands more fine-grained workflow design from the upstream, and training strengthens upstream agents to better satisfy that demand.}

\begin{wrapfigure}[8]{r}{0.5\textwidth}
    \vspace{-60pt}
    \centering
    \includegraphics[width=0.48\textwidth]{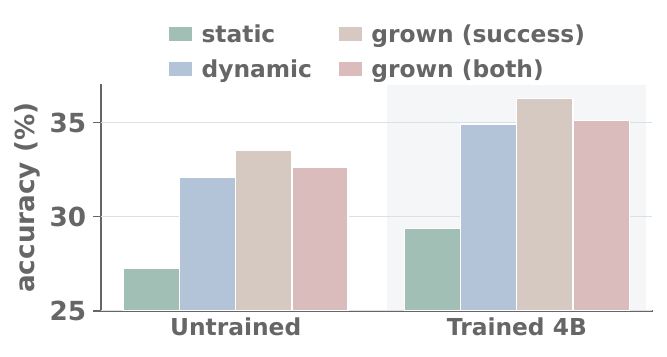}

    \vspace{-10pt}

    \caption{\textbf{Pool Dynamics.} A \textit{static} pool stays fixed, a \textit{dynamic} pool expands on demand, and a \textit{grown} pool also carries components distilled from past experiences (\S\ref{subsec:method:wire}).}
    \label{fig:exp_pool_dynamics}

    \vspace{-10pt}
\end{wrapfigure}

\textbf{Growing the pool from successful experiences adds the most.} As shown in Fig.~\ref{fig:exp_pool_dynamics}, a pool that grows on demand gains $+4.83\%$ over a pool held fixed. This also gives the largest effect among all the single design choices in this ablation study. Distilling past experiences into the pool adds a further $+1.44\%$ when only successful ones are kept, against $+0.53\%$ when both successful and failed ones are kept. The performance ranking among the four pool dynamics modes holds before and after optimization, with gains up to $+2.83\%$ from training. \textit{The workflow therefore learns from its own experience, and learns most from the experiences that succeed.}

\section{Related Work}
\label{sec:main_related_work}

We summarize four lines of work below and extend each in \S\ref{sec:related_work}.

\textbf{Workflow Construction and Optimization.}
Early multi-agent frameworks coordinate agents through manually specified roles and procedures~\citep{hong2024metagpt,wu2023autogen}.
Later methods automate this design while keeping every model fixed, searching for a workflow written as code~\citep{zhang2025aflow,hu2025adas} or generating a graph for each task and revising it during execution~\citep{niu2025flowmodularizedagenticworkflow}, and a further line trains the \textit{generator} itself from execution outcomes~\citep{zhang2025maas,nie2025weakforstrongtrainingweakmetaagent,nielsen2026learningorchestrateagents}.
A separate line improves the workflow without updating any weights, inducing reusable routines from past experience~\citep{wang2024agentworkflowmemory}, evolving the context a model conditions on~\citep{zhang2026ace}, or replacing reinforcement learning with reflective prompt evolution~\citep{agrawal2026gepa}.
\textit{However, the models that solve the workflow's subtasks remain fixed, limiting how much the workflow as a whole can improve}.

\textbf{Multi-Agent Training and Co-Evolution.}
Reinforcement learning increasingly optimizes several interacting agents, yet within an interaction structure \textit{fixed in advance}, such as a discussion among agents that play one role~\citep{park2025maporl} or a workflow designed once per task domain~\citep{zhao2026strongermas}.
Co-evolution likewise fixes how the models interact, pairing a challenger that proposes problems with a solver that answers them~\citep{huang2026rzero}, or scoring proposals in a multi-round discussion~\citep{xue2026comas}.
Other works train each agent separately, in its own reinforcement learning with a role-specific dataset and reward~\citep{guo2026anagentenhancingscientifictable,motwani2025malt}.
\textit{However, agents then learn in isolation or in one fixed role, rather than over the workflows constructed for each task}.

\textbf{Credit Assignment and Failure Attribution.}
Credit finer than a single outcome typically comes from \textit{learned models}, such as a process reward model trained alongside the policy~\citep{liu2025agenticreinforcementlearningimplicit} or a local reward for each component, estimated by re-executing what follows it~\citep{wu2026optimas}.
Failure attribution instead names the responsible agent and step with LLM judges~\citep{zhang2025whowhen}, an attributor trained on labeled failures~\citep{zhang2025agentracer}, or re-execution with edited messages~\citep{ma2026dover}, yet identifies the decisive step in only about a third of failed runs~\citep{chen2026traceelephant}.
Credit that adds \textit{no model or rollout} compares actions taken from states that recur across a single agent's rollouts of the same task~\citep{feng2025gigpo,cheng2026graphgpo}.
\textit{However, such attribution costs models, labels, or executions beyond the workflow run itself, and it can still misattribute where a failure originates}.

\textbf{Evaluation of Multi-Agent Workflows.}
Methods that construct or optimize workflows are commonly evaluated on single-question math, function-level code, and short question answering~\citep{zhang2025aflow,zhang2025maas,nie2025weakforstrongtrainingweakmetaagent}.
On such data, automatically designed systems improve on a self-consistent single agent by at most a few points~\citep{singh2025grammarsearch}, and majority voting alone accounts for most of the gains attributed to multi-agent debate~\citep{choi2025debatevoteyieldsbetter}.
Training on it can also collapse a system into \textit{one dominant agent}~\citep{zhang2025lazyagents}, and failure analyses find the reported gains often minimal~\citep{cemri2025multiagentllmsystemsfail}.
\textit{However, such data can neither reveal nor optimize what a workflow adds beyond a single agent}.

\section{Conclusions}
\label{sec:conclusions}

In this work, we introduce \ours, which leverages the workflow as harness (\S\ref{subsec:method:wire}) to provide the optimization signal without additional annotations: one signal that evaluates a workflow, trains any role that builds or executes it, and evolves a reusable prior at test time (\S\ref{subsec:prelim:harness}, \ref{subsec:method:objective}). A hierarchical, structure-aware reward (\S\ref{subsec:method:reward}) turns the sparse outcome of a workflow into role-level feedback, with role-specific credit derived from the execution trace (\S\ref{subsec:method:credit}). \ours enables train-time optimization in four modes: (1) single role self-evolution, (2) agent-skill co-evolution, (3) upstream-downstream co-evolution, and (4) two or more roles co-evolve; and test-time optimization via meta distillation (\S\ref{subsec:method:evolve}). To supply reliable workflow-level evaluation, we further propose \oursdata to convert single agent level datasets into workflow-level tasks through three hardening strategies at two levels, yielding the $44$ arms our experiments span (\S\ref{appendix:sec:data_adaptation}).
Across those arms, \ours outperforms the single-agent baselines by $\geq+11.39\%$, adds a further $+2.58\%$ through meta distillation and $+5.03\%$ through reinforcement optimization (\S\ref{subsec:results:main}). These gains hold across RL algorithms, workflow topologies, pool granularities, and pool dynamics (\S\ref{subsec:results:ablation}), and transfer to backbones the harness never trains (\S\ref{appendix:subsec:results:generalization}).

\clearpage

\bibliography{iclr2027_conference}
\bibliographystyle{iclr2027_conference}

\clearpage
\appendix

\section{Extended Discussion of Related Work}
\label{sec:related_work}

\subsection{Workflow Construction and Optimization}
\label{subsec:related:generation}

Early multi-agent frameworks coordinate agents through manually specified roles, procedures, and conversation patterns~\citep{hong2024metagpt,wu2023autogen}, and recent visual document understanding systems run agent pipelines designed by hand for their domain on fixed, prompted backbones~\citep{zhu2025doclenstoolaugmentedmultiagent,jin2026slideagent}.
Other methods automate this design while keeping every model fixed, searching on each benchmark's validation data for a single workflow or agent design written as code~\citep{zhang2025aflow,hu2025adas}, or generating a workflow graph for each task and revising it during execution~\citep{niu2025flowmodularizedagenticworkflow}.
Methods that train the workflow generator use supervised fine-tuning on query-specific multi-agent systems~\citep{ye2025masgpt} or learn from execution outcomes, training lightweight controllers that select operators~\citep{zhang2025maas} or communication edges~\citep{zhang2025gdesigner} for fixed LLMs, a weak meta-agent that designs workflows for strong executors~\citep{nie2025weakforstrongtrainingweakmetaagent}, or orchestrators that assign subtasks to fixed worker models~\citep{nielsen2026learningorchestrateagents} or select the next agent step by step~\citep{dang2025puppeteer}.
Such training also extends to several meta-agents that generate, configure, and rectify a multi-agent system whose agents run on a fixed model pool~\citep{wang2025mas2}.
A separate line improves the workflow without updating weights, leveraging reusable routines from past experience~\citep{wang2024agentworkflowmemory}, evolving the context a model conditions on~\citep{zhang2026ace}, or replacing reinforcement learning with reflective prompt evolution~\citep{agrawal2026gepa}.
\textit{However, in all these methods, the models that solve the workflow's subtasks remain fixed, which limits how much the workflow as a whole can improve}.

\subsection{Multi-Agent Training and Co-Evolution}
\label{subsec:related:multirole}

Reinforcement learning increasingly optimizes multiple interacting LLM agents, but within an interaction structure fixed in advance for all task instances, such as a multi-turn discussion among agents that play the same role~\citep{park2025maporl}, a meta-thinking agent that guides a reasoning agent~\citep{wan2025rema}, a workflow designed once per task domain~\citep{zhao2026strongermas}, or a document-understanding pipeline of planning, execution, judgment, and answering agents~\citep{yu2025visualdocumentunderstandingreasoning}.
Co-evolution methods likewise fix how the models interact, pairing a challenger that proposes problems with a solver that answers them~\citep{huang2026rzero}, or letting agents drawn at random propose, critique, and score solutions in a multi-round discussion whose scores serve as their rewards~\citep{xue2026comas}.
Recursive agent training lets the task decomposition vary per instance, but trains a single policy as a root agent and as the copies of itself to which it delegates, all in the same role~\citep{gandhi2026rao}.
Other works train each agent separately, optimizing it in its own reinforcement learning with a role-specific dataset and reward~\citep{guo2026anagentenhancingscientifictable}, or fine-tuning it to draft, verify, or refine an answer on trajectories sampled once from the untrained model~\citep{motwani2025malt}.
\textit{However, these agents learn either in isolation or only within the one structure or role they are trained in, which leaves them unprepared for tasks whose outcome depends on different agents working together in different ways}.

\subsection{Credit Assignment and Failure Attribution}
\label{subsec:related:credit}

Credit finer than a single outcome often comes from learned models, such as a process reward model trained alongside a single agent's policy to reward each action~\citep{liu2025agenticreinforcementlearningimplicit}, or a local reward for each component of a compound system, learned from outcomes estimated by re-executing the components downstream of it~\citep{wu2026optimas}.
Other methods credit the meta-agents that design a multi-agent system by averaging the outcomes of executions branched at each of their decisions~\citep{wang2025mas2}.
Failure attribution for multi-agent systems names the responsible agent and step with LLM judges~\citep{zhang2025whowhen} or with an attributor trained on failures labeled through counterfactual replay and fault injection~\citep{zhang2025agentracer}, or tests failure hypotheses by re-executing the system with edited messages or plans~\citep{ma2026dover}.
Even with full execution traces, such attributors identify the decisive failure step in only about a third of failed runs~\citep{chen2026traceelephant}.
Credit that adds no model or rollout instead compares actions taken from states that recur across a single agent's rollouts of the same task, and falls back to trajectory-level credit when states do not recur~\citep{feng2025gigpo,cheng2026graphgpo}.
\textit{However, such credit either costs extra models, labels, or executions, or holds only where a single agent revisits the same states, and it can still misattribute failures. Consequently, agents that share one outcome lack a cheap and reliable signal of their own mistakes}.

\subsection{Evaluation of Multi-Agent Workflows}
\label{subsec:related:benchmarks}

Methods that construct or optimize multi-agent workflows are commonly evaluated on single-question math, function-level code, and short question answering~\citep{zhang2025aflow,zhang2025maas,nie2025weakforstrongtrainingweakmetaagent}.
On such data, automatically designed multi-agent systems improve on a self-consistent single agent by at most a few points~\citep{singh2025grammarsearch}, many reported successes of efficiency methods for multi-agent systems appear in settings where a single agent already preserves strong performance~\citep{zhang2026rethinkingevaluationefficiencymethods}, and majority voting alone accounts for most of the gains attributed to multi-agent debate~\citep{choi2025debatevoteyieldsbetter}.
When multiple agents are trained on single-question math, one agent comes to dominate while the other contributes little, collapsing the system into a single agent~\citep{zhang2025lazyagents}.
Final answer accuracy can also remain high while the interactions among agents diverge from their intended behavior~\citep{so2026superficial}, and failure analyses find the gains of multi-agent systems on popular benchmarks often minimal~\citep{cemri2025multiagentllmsystemsfail}.
\textit{However, as such data can be solved without coordinating agents, evaluating on it cannot reliably assess whether a workflow's structure helps, and training on it provides little signal for learning beyond a single agent's effort}.

\section{Limitations}
\label{sec:limitations}

By proposing \ours, there are also limitations that confine the scope of what we demonstrate.
First, our reinforcement learning experiments are mainly evaluated on two open models of a single family, \texttt{Qwen3.5-4B} and \texttt{Qwen3.5-9B} (\S\ref{sec:exps}). \ours leverages the same signal whichever model powers a role, and the meta-distillation improves all four backbones we evaluate, including the two it never optimizes (\S\ref{appendix:subsec:results:generalization}). Nevertheless, we aim to incorporate more model families into our evaluation to strengthen our work.
Second, our evaluation spans $44$ arms over $12$ datasets and $7$ domains (\S\ref{appendix:subsec:data_adaptation:stats}), each a task whose inputs are fixed once the task begins. In future work, we plan to train roles at a larger scale, and to extend \ours to dynamic environments, where the inputs a workflow reads and the components it needs change while the workflow runs.

\section{Preliminary Study: The Gap in Existing Benchmarks}
\label{appendix:sec:preliminary}

A multi-agent workflow is only warranted on a task that a single agent cannot already solve. We therefore ask two questions: what existing benchmarks demand of a system, whether a single agent or a workflow, and how much a workflow gains on them. We answer the first by surveying the evaluations that published workflow generation methods use, and the second by a controlled comparison in which the workflow and the single agent both run the untrained backbone \texttt{Qwen3-VL-8B}.

\begin{figure}[h]
    \centering
    \includegraphics[width=1.0\textwidth]{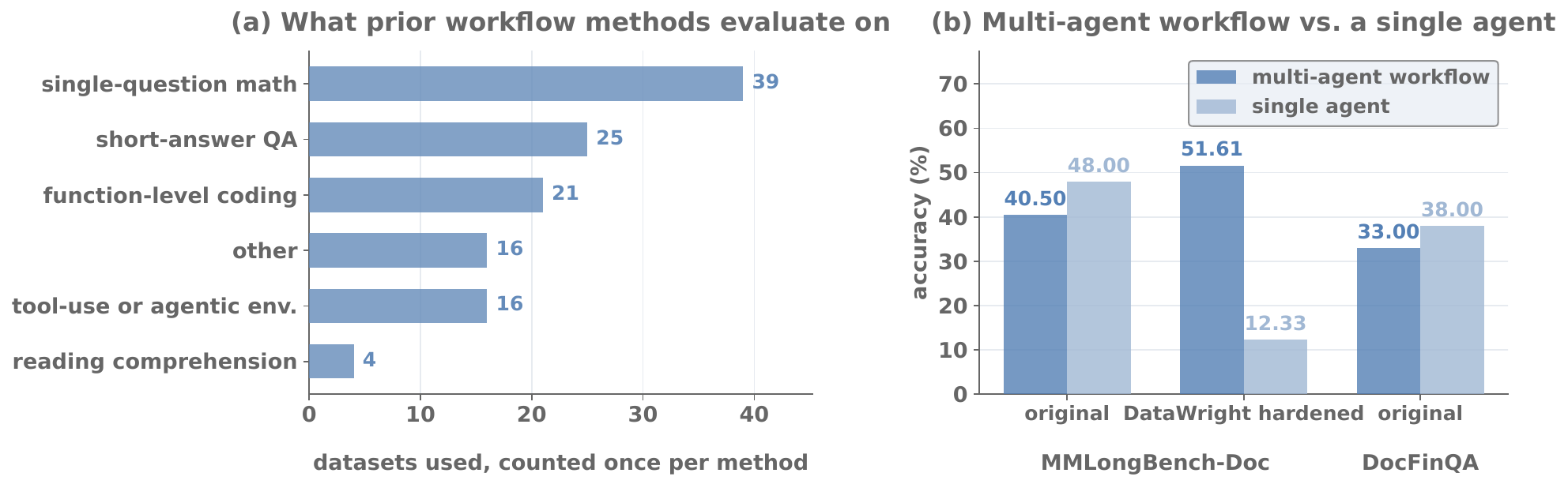}
    \caption{\textbf{Preliminary Study.} (a) Among the $121$ datasets that the $20$ workflow generation methods evaluate on: $85$ of the $121$ are single-question math, short-answer QA, or function-level coding, each of which a single competent agent can answer in one pass. (b) Comparing multi-agent workflows against a single agent given the same compute budget: workflows underperform the single agent on the original datasets, yet gain $\uparrow39.28$ on \oursdata-hardened dataset.}
    \label{fig:preliminary_study}
\end{figure}

\textbf{What Existing Benchmarks Demand.}
We collect every dataset that each of $20$ workflow generation methods published at top-tier venues evaluates on~\citep{zhang2025aflow,hu2025adas,zhang2025maas,zhuge2024gptswarm,shang2024agentsquare,nie2025weakforstrongtrainingweakmetaagent,ye2025masgpt,wang2025mas2,dang2025puppeteer,ke2026masorchestra,nielsen2026learningorchestrateagents,su2025toolorchestra,zhao2026strongermas,zhang2025gdesigner,niu2025flowmodularizedagenticworkflow,zhang2025metaagent,zhou2026multiagentdesign,wang2025dyflow,hu2025owl,singh2025grammarsearch}, which gives $121$ datasets in total, counting a dataset once for each method that uses it. We classify these $121$ datasets into six categories according to what a system needs to do to answer each question (Fig.~\ref{fig:preliminary_study}a): single-question math ($39$), short-answer QA ($25$), function-level coding ($21$), tool use or agentic environments ($16$), reading comprehension ($4$), and other ($16$). The first three categories account for $85$ of the $121$, and a single competent agent is able to answer their questions in one pass~\footnote{These counts describe the $20$ methods we survey, and we do not extrapolate them to the field.}.

\textbf{How Much a Workflow Gains on Them.}
Leveraging \texttt{Qwen3-VL-8B} as the backbone, we compare LLM-generated multi-agent workflow against a single agent on two datasets of this kind (MMLongBench-Doc~\citep{ma2024mmlongbenchdoc} and DocFinQA~\citep{reddy2024docfinqa}). We give the workflow and the single agent the same compute budget, so that any difference in accuracy is attributable to how the computation is organized rather than to how much of it each system is allowed (Fig.~\ref{fig:preliminary_study}b). On the original MMLongBench-Doc (200-sample subset) the workflow reaches $40.50\%$ and the single agent $48.00\%$. On the original DocFinQA (200-sample subset) the workflow reaches $33.00\%$ and the single agent $38.00\%$. Neither dataset rewards multi-agent coordination. Each sample presents one document and one question, so a system reads through the document and extracts one answer, and no decision in the task requires ability or coordination beyond a single agent. The outcome therefore reflects what these benchmarks ask for, not a limitation of the workflow.

\textbf{Hardening Changes the Outcome.}
On MMLongBench-Doc hardened by \oursdata, the same workflow reaches $51.61\%$ against the single agent's $12.33\%$ over $150$ test samples, an improvement of $39.28$ points. Through hardening (\S\ref{appendix:subsec:data_adaptation:strategies}), the evidence a system needs to examine exceeds one effective context window. Therefore, the task requires deciding which input to read in what order, and spending the same number of turns on a single input no longer suffices. This is the condition under which the structure and coordination a workflow adds carry value~\footnote{We observe this reversal in our preliminary study, so we analyze it as an existence result that motivates \oursdata (\S\ref{appendix:sec:data_adaptation}), and not as a rate at which hardening reverses an outcome.}.

\section{\oursdata: Workflow-Level Data Hardening}
\label{appendix:sec:data_adaptation}

Datasets that a single agent can already handle can neither reveal nor optimize what a workflow adds beyond it (\S\ref{appendix:sec:preliminary}). To this end, we propose \oursdata to convert existing datasets into workflow-level tasks. Particularly, it builds each task by a rule-based \textit{hardening strategy} at a \textit{hardening level} $\ell$ (\S\ref{appendix:subsec:data_adaptation:strategies}), adaptively selects the strategies each dataset supports (\S\ref{appendix:subsec:data_adaptation:selection}), and filters the data with quality control (\S\ref{appendix:subsec:data_adaptation:qc}). Each hardened task is graded as the mean of its sub-question scores, and the accuracy of each hardened dataset is calculated as the mean accuracy over its test tasks (\S\ref{appendix:subsec:data_adaptation:eval_metrics}). Alg.~\ref{alg:data_adaptation} summarizes the workflow-level data hardening procedure of \oursdata. We also summarize our data statistics in \S\ref{appendix:subsec:data_adaptation:stats}.

\subsection{Hardening Strategies}
\label{appendix:subsec:data_adaptation:strategies}

A hardening strategy converts samples of an existing dataset, each a question paired with the input it asks about, such as a document, a slide deck, a chart, \textit{etc.}, into harder tasks by its rule, without any model in the loop. Solving a hardened task draws on the capabilities that a workflow brings beyond a single agent, including decomposing the task into sub-questions, routing each sub-question to a suitable input and agent, localizing the evidence that answers it, and aggregating the answers. Every strategy calls on these capabilities together, while each emphasizes a different one by its construction.

\textbf{Shared} bundles $\ell$ questions about the same input into one task. Solving it mainly exercises \textit{decomposition}: the task splits into sub-questions that can be assigned and answered separately.

\textbf{Paired} bundles $\ell$ questions, each with its own distinct input, into one task. Solving it mainly exercises \textit{routing}: each sub-question is matched to the input that answers it.

\textbf{Decoy} keeps one question and places its input among $(\ell-1)$ decoy inputs drawn from the same dataset. Solving it mainly exercises \textit{localization}: finding the single input that answers the question. Every input serves as a decoy a near-equal number of times, never for its own question.

\textbf{Hardening level.} The level $\ell$ sets the number of questions per task for \textit{shared} and \textit{paired}, and the number of inputs per task for \textit{decoy}. Each original question enters at most one task, and every dataset is split into training and test sets before hardening. Therefore, no question appears on both sides at any level. Accordingly, a hardened task's score is the mean of its sub-question scores (\S\ref{appendix:subsec:data_adaptation:eval_metrics}).

\begin{algorithm}[t]
\caption{\textbf{\oursdata: Workflow-Level Data Hardening.} A sample $(q, z, y^\star)$ is a question $q$ with its input $z$ and reference answer $y^\star$; $\mathcal{B}_z$ denotes the samples of $\mathcal{B}$ on input $z$, and $|x|$ the number of original samples in task $x$. \textsc{Group} partitions samples into disjoint tasks of $\ell$ questions, and \textsc{Decoy} draws $\ell-1$ inputs other than $z$, using every input a near-equal number of times.}
\label{alg:data_adaptation}
\small
\begin{algorithmic}[1]
\Function{DataWright}{$\mathcal{B}, \ell$}
  \State $\mathcal{B} \gets \{(q, z, y^\star) \in \mathcal{B} \mid \textsc{Valid}(q, z, y^\star)\}$ \Comment{\S\ref{appendix:subsec:data_adaptation:qc}}
  \State $(\mathcal{B}^{\mathrm{train}}, \mathcal{B}^{\mathrm{test}}) \gets \textsc{Split}(\mathcal{B})$
  \For{$\sigma \in \{\sigma' \mid \textsc{Supports}(\sigma', \mathcal{B})\}$} \Comment{\S\ref{appendix:subsec:data_adaptation:selection}}
    \State $\mathcal{T}^{\mathrm{train}}_{\sigma} \gets \textsc{Harden}(\mathcal{B}^{\mathrm{train}}, \sigma, \ell)$,\quad $\mathcal{T}^{\mathrm{test}}_{\sigma} \gets \textsc{Harden}(\mathcal{B}^{\mathrm{test}}, \sigma, \ell)$
    \If{$\sum_{x \in \mathcal{T}^{\mathrm{test}}_{\sigma}} |x| < 100$}
      \State $\mathcal{T}^{\mathrm{train}}_{\sigma}, \mathcal{T}^{\mathrm{test}}_{\sigma} \gets \varnothing$
    \EndIf
  \EndFor
  \State \Return $\{(\mathcal{T}^{\mathrm{train}}_{\sigma}, \mathcal{T}^{\mathrm{test}}_{\sigma})\}_{\sigma}$
\EndFunction
\Statex
\Function{Harden}{$\mathcal{B}, \sigma, \ell$} \Comment{\S\ref{appendix:subsec:data_adaptation:strategies}}
  \If{$\sigma = \text{Shared}$}
    \State \Return $\bigcup_{z} \textsc{Group}(\mathcal{B}_z, \ell)$
  \ElsIf{$\sigma = \text{Paired}$}
    \State \Return $\textsc{Group}(\mathcal{B}, \ell)$ \Comment{distinct $z$ within a task}
  \Else
    \State \Return $\{(q, \textsc{Shuffle}(\{z\} \cup \textsc{Decoy}(z, \ell - 1)), y^\star) \mid (q, z, y^\star) \in \mathcal{B}\}$
  \EndIf
\EndFunction
\end{algorithmic}
\end{algorithm}

\subsection{Adaptive Strategy Selection}
\label{appendix:subsec:data_adaptation:selection}

Not every dataset supports every hardening strategy, so \oursdata selects the suitable strategies for each dataset from what its samples allow (\textsc{Supports} in Alg.~\ref{alg:data_adaptation}). Each strategy has one prerequisite.

\textbf{Shared} requires inputs that each carry several independent questions. It is therefore unavailable for (1) datasets that pair every input with a single question, such as one question per coding problem (\textit{e.g.}, TACO~\citep{li2023taco}), per image (\textit{e.g.}, MMMU-Pro~\citep{yue2025mmmupro}), per chart (\textit{e.g.}, ChartCoder~\citep{zhao2025chartcoder}), \textit{etc.}, and for (2) datasets whose questions about one input are steps of the same problem (\textit{e.g.}, We-Math~\citep{qiao2024wemath}), where one step reveals another's answer.

\textbf{Paired} requires only questions with distinct inputs, which every dataset provides.

\textbf{Decoy} requires decoys that cannot answer the question, which holds only when the question identifies its own input, as questions about a specific document, slide deck, or financial filing do. For charts, images, and coding problems, a decoy of the same kind can match everything the question states, which would make the task ambiguous rather than harder, so \oursdata does not apply this strategy to them.

\textbf{Sample-size requirement.} A hardening level of a strategy is kept for a dataset only if its test set at that level holds at least 100 original samples, so that every accuracy score rests on enough samples to give a valid evaluation. Tab.~\ref{tab:data_stats} summarizes the strategies and levels selected for each dataset.

\subsection{Quality Control}
\label{appendix:subsec:data_adaptation:qc}

Existing datasets contain defective samples, such as a question that its input cannot answer, an input that does not match its question, or a reference answer that the input contradicts.
\oursdata therefore validates the samples of each dataset before hardening (\textsc{Valid} in Alg.~\ref{alg:data_adaptation}), mainly by quality filtering and automatic checks.

\textbf{Quality filtering.}
On the training data, GPT-5.4 serves as an LLM judge: it examines each sample together with its reference answer and flags the answers it finds wrong. The flagged samples are removed, 203 of the 21,570 training samples (1.9\%). For example, on DocFinQA, the judge removes 79 unanswerable questions, and on TACO it removes 36 problems whose statement relies on a figure that the dataset no longer provides.

\textbf{Automatic checks.}
For code generation datasets, we implement automatic checks by running the reference solution of each sample, keeping the samples that can pass their own tests. For example, a TACO problem requires at least 10 tests, so that an incorrect code answer cannot pass by chance.

\subsection{Evaluation Metrics}
\label{appendix:subsec:data_adaptation:eval_metrics}

Hardening changes how hard a task is, not how its answers are graded: every sub-question keeps the metric of its own answer type, and a task's score is the mean over its sub-questions. A hardened task $x$ holds $|x|$ sub-questions, the $i$-th carrying the reference answer $y^\star_i$ and receiving the answer $y_i$ that a workflow returns: $|x| = \ell$ under \textit{shared} and \textit{paired}, which bundle $\ell$ questions per task, and $|x| = 1$ under \textit{decoy}, whose level counts inputs rather than questions. With $s(\cdot, \cdot) \in [0, 1]$ denoting the metric of that sub-question's answer type, a task's score and a dataset's accuracy over its test tasks are:
\begin{equation}
S(x) = \frac{1}{|x|} \sum_{i=1}^{|x|} s(y_i, y^\star_i),
\qquad
\mathrm{Acc}(\mathcal{T}^{\mathrm{test}}_{\sigma}) = \frac{1}{|\mathcal{T}^{\mathrm{test}}_{\sigma}|} \sum_{x \in \mathcal{T}^{\mathrm{test}}_{\sigma}} S(x)
\label{eq:metric_task}
\end{equation}
where a sub-question the workflow leaves unanswered scores $s = 0$. Tab.~\ref{tab:metrics} summarizes the evaluation metrics for different types of tasks \ours cover. We also show the task types of each dataset in Tab.~\ref{tab:data_stats}.

\textbf{Multiple choice and lists.}
Let $Y$ and $Y^\star$ be the sets of options, or of list items, that the answer and the reference name. The score is the overlap of the two sets against the larger of them:
\begin{equation}
s_{\mathrm{set}}(Y, Y^\star) = \frac{|Y \cap Y^\star|}{\max\left(|Y^\star|, |Y|\right)}
\label{eq:metric_choice}
\end{equation}
which reaches 1 only when the two sets agree. Extra names are penalized, just as missing ones are, so a workflow cannot gain by naming every option. A single-answer question is the case $|Y^\star| = 1$, where Eq.~\ref{eq:metric_choice} is an exact match.

\textbf{Numeric.}
A numeric answer is first normalized: thousands separators and currency symbols are removed, units are mapped to a canonical name, and a percentage and its fraction become one quantity. It is then compared at the precision the reference itself states, with $\mathrm{round}_{y^\star}(\cdot)$ rounding to that precision:
\begin{equation}
s_{\mathrm{num}}(y, y^\star) = \one\!\left[\mathrm{round}_{y^\star}(y) = \mathrm{round}_{y^\star}(y^\star)\right]
\label{eq:metric_numeric}
\end{equation}
No relative tolerance is allowed, which is stricter than some of these benchmarks.

\textbf{Open-ended.}
Let $\nu(\cdot)$ normalize case, punctuation, and spacing, and $\mathrm{tok}(\cdot)$ be the multiset of answer tokens. An exact match after normalization scores 1, and otherwise the score is the token-level F1 with the reference:
\begin{equation}
s_{\mathrm{text}}(y, y^\star) =
\begin{cases}
1, & \nu(y) = \nu(y^\star), \\[4pt]
\dfrac{2 \left|\mathrm{tok}(y) \cap \mathrm{tok}(y^\star)\right|}{\left|\mathrm{tok}(y)\right| + \left|\mathrm{tok}(y^\star)\right|}, & \text{otherwise}
\end{cases}
\label{eq:metric_text}
\end{equation}
so a sub-answer that overlaps its reference without matching it earns partial credit. The second case applies inside \textit{shared} and \textit{paired} tasks. A \textit{decoy} task holding a single question is scored by the first case alone.

\textbf{Code generation.}
A code is graded by running it, never by comparing its text to a reference:
\begin{equation}
s_{\mathrm{code}}(y) = \one\!\left[\textsc{Check}(y)\right]
\label{eq:metric_code}
\end{equation}
where \textsc{Check} is a strict execution-based evaluation test: every unit test of the problem passes, the benchmark's own checker test accepts the code, or, for a dataset that provides no evaluation protocol, the code executes without errors and the code-rendered output is accurate against the reference answer.


\textbf{LLM as Judge.}
An answer that no rule can decide is graded by an LLM judge that evaluates the answer directly against the reference. As such, we define an LLM judge as a grader that receives the reference answer and the predicted answer, and evaluates with one integer on a $\{0, \dots, L\}$ scale whose every level its prompt pre-defines:
\begin{equation}
s_{\mathrm{judge}}(y, y^\star) = \frac{1}{L}\,\textsc{Judge}\!\left(y, y^\star\right)
\label{eq:metric_judge}
\end{equation}
where the judge returns an integer in $\{0, \dots, L\}$ and $L$ is the top of the scale its prompt defines. The prompt and the normalization thus stay in step. Each judge makes exactly one judgment, which keeps an answer self-consistent: a verdict paired with a separate similarity admits a pair that disagrees with itself, while anchored levels calibrate. A judge reads text or images, whichever the answer is made of, and a dataset that needs several judges gets one per aspect of the same answer, which the metric averages into that task's score.

\begin{figure}[t]
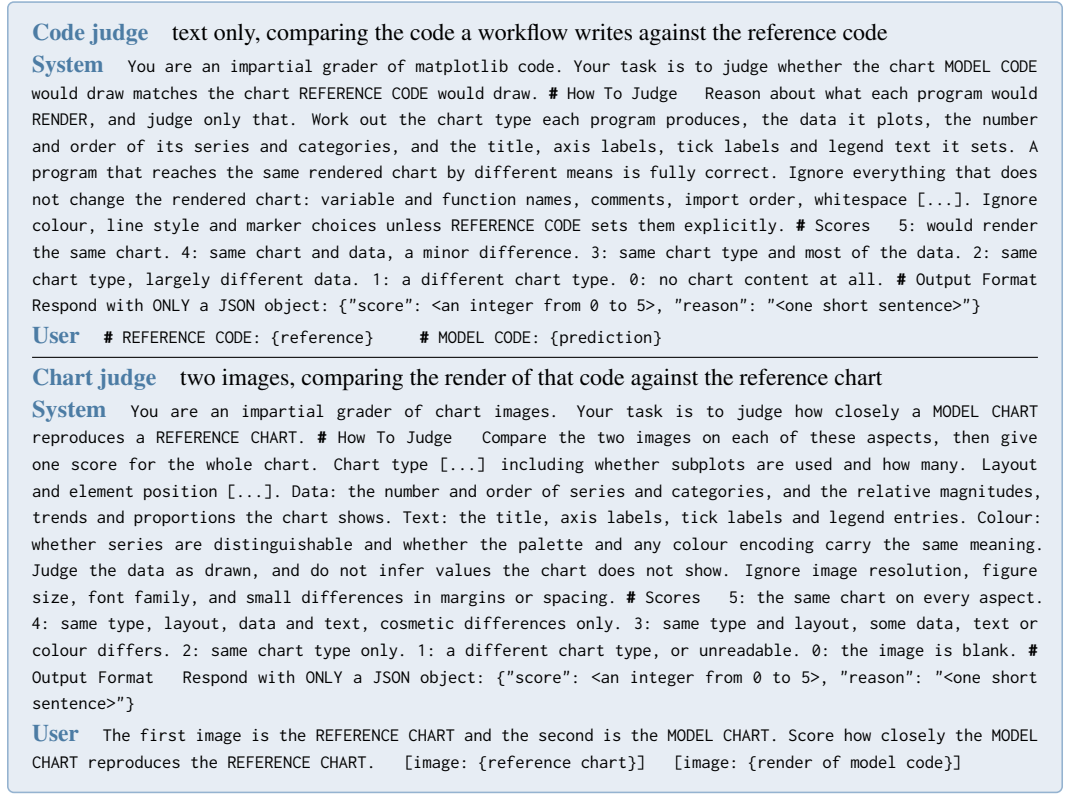

\centering
\small
\begin{tcolorbox}[colback=bboxface, colframe=bboxedge, boxrule=0.6pt, arc=2pt, left=6pt, right=6pt, top=5pt, bottom=5pt]

{\bfseries\textcolor{bboxtag}{Code judge}}\quad text only, comparing the code a workflow writes against the reference code

\vspace{2pt}

{\bfseries\textcolor{bboxtag}{System}}\quad {\ttfamily\scriptsize You are an impartial grader of matplotlib code. Your task is to judge whether the chart MODEL CODE would draw matches the chart REFERENCE CODE would draw. \textbf{\#} How To Judge \quad Reason about what each program would RENDER, and judge only that. Work out the chart type each program produces, the data it plots, the number and order of its series and categories, and the title, axis labels, tick labels and legend text it sets. A program that reaches the same rendered chart by different means is fully correct. Ignore everything that does not change the rendered chart: variable and function names, comments, import order, whitespace \texttt{[...]}. Ignore colour, line style and marker choices unless REFERENCE CODE sets them explicitly. \textbf{\#} Scores \quad 5: would render the same chart. 4: same chart and data, a minor difference. 3: same chart type and most of the data. 2: same chart type, largely different data. 1: a different chart type. 0: no chart content at all. \textbf{\#} Output Format \quad Respond with ONLY a JSON object: \{"score": <an integer from 0 to 5>, "reason": "<one short sentence>"\}}

\vspace{2pt}

{\bfseries\textcolor{bboxtag}{User}}\quad {\ttfamily\scriptsize \textbf{\#} REFERENCE CODE: \{reference\} \quad\quad \textbf{\#} MODEL CODE: \{prediction\}}

\vspace{4pt}\hrule\vspace{4pt}

{\bfseries\textcolor{bboxtag}{Chart judge}}\quad two images, comparing the render of that code against the reference chart

\vspace{2pt}

{\bfseries\textcolor{bboxtag}{System}}\quad {\ttfamily\scriptsize You are an impartial grader of chart images. Your task is to judge how closely a MODEL CHART reproduces a REFERENCE CHART. \textbf{\#} How To Judge \quad Compare the two images on each of these aspects, then give one score for the whole chart. Chart type \texttt{[...]} including whether subplots are used and how many. Layout and element position \texttt{[...]}. Data: the number and order of series and categories, and the relative magnitudes, trends and proportions the chart shows. Text: the title, axis labels, tick labels and legend entries. Colour: whether series are distinguishable and whether the palette and any colour encoding carry the same meaning. Judge the data as drawn, and do not infer values the chart does not show. Ignore image resolution, figure size, font family, and small differences in margins or spacing. \textbf{\#} Scores \quad 5: the same chart on every aspect. 4: same type, layout, data and text, cosmetic differences only. 3: same type and layout, some data, text or colour differs. 2: same chart type only. 1: a different chart type, or unreadable. 0: the image is blank. \textbf{\#} Output Format \quad Respond with ONLY a JSON object: \{"score": <an integer from 0 to 5>, "reason": "<one short sentence>"\}}

\vspace{2pt}

{\bfseries\textcolor{bboxtag}{User}}\quad {\ttfamily\scriptsize The first image is the REFERENCE CHART and the second is the MODEL CHART. Score how closely the MODEL CHART reproduces the REFERENCE CHART. \quad [image: \{reference chart\}] \quad [image: \{render of model code\}]}

\end{tcolorbox}

\vspace{-10pt}

\caption{\textbf{LLM Judge Prompts.} Each judge prompt carries its system instruction and its user message (\S\ref{appendix:subsec:data_adaptation:eval_metrics}), trimmed with \texttt{[...]}. Scores are graded to $0$ only when code execution failed (Eq.~\ref{eq:metric_chart}).}
\label{fig:appendix:llm_judge_prompt}
\end{figure}

For ChartCoder, we propose its evaluation as the mean of two LLM judged scores gated by execution: two judges at $L{=}5$ (Fig.~\ref{fig:appendix:llm_judge_prompt}) alongside the execution term of Eq.~\ref{eq:metric_code}. The first LLM judge evaluates code by assessing the predicted code against the reference code. The second LLM judge evaluates the predicted chart rendered by the predicted code against the reference chart as two images. When execution fails, the score becomes $0$:
\begin{equation}
\begin{aligned}
s^{\mathrm{code}}_{\mathrm{chart}}(y, y^\star) &= \tfrac{1}{5}\,\one\!\left[\textsc{Exec}(y)\right] \textsc{Judge}_{\mathrm{code}}\!\left(y, y^\star\right), \\[2pt]
s^{\mathrm{render}}_{\mathrm{chart}}(y, y^\star) &= \tfrac{1}{5}\,\one\!\left[\textsc{Exec}(y)\right] \textsc{Judge}_{\mathrm{render}}\!\left(\textsc{Render}(y), \textsc{Render}(y^\star)\right), \\[2pt]
s_{\mathrm{chart}}(y, y^\star) &= \tfrac{1}{2}\left(s^{\mathrm{code}}_{\mathrm{chart}}(y, y^\star) + s^{\mathrm{render}}_{\mathrm{chart}}(y, y^\star)\right)
\end{aligned}
\label{eq:metric_chart}
\end{equation}
where $\textsc{Exec}$ marks a code that executes successfully, and $\textsc{Render}$ renders one to an image. Both judges score $0$ unless the code executes without error. The two judges cover two different aspects of one answer: one evaluating the code from its source and the other the image that code renders, and their mean is the accuracy we present for each ChartCoder task. We use \texttt{GPT-5.6-luna} as the judge (\S\ref{sec:exps}). ChartCoder is the one dataset that requires a judge, so every ChartCoder number in our experiments is a judge score, and every other dataset carries the rule-based metric of its answer type defined above.

\begin{table}[H]

\vspace{0pt}

\centering
\caption{\textbf{Evaluation Metrics.} We summarize the evaluation metrics of different types of tasks covered by \oursdata (\S\ref{appendix:subsec:data_adaptation:eval_metrics}).}
\label{tab:metrics}
\small
\renewcommand{\arraystretch}{1.2}
\newcommand{\blockgap}{\noalign{\vskip 2pt}}

\begin{tabularx}{\linewidth}{@{}p{0.32\linewidth}Xp{0.10\linewidth}@{}}
\toprule
\multicolumn{1}{@{}c}{\textbf{Task Type}} & \multicolumn{1}{c}{\textbf{Evaluation Metric}} & \multicolumn{1}{c@{}}{\textbf{Definition}} \\
\midrule
\rowcolor{blue_bg}
\multicolumn{3}{@{}c@{}}{\textcolor{blue_text}{\textbf{\textit{Question Answering}}}} \\
\blockgap
multiple choices, one answer & the chosen option equals the reference option & Eq.~\ref{eq:metric_choice} \\
multiple choices, multiple answers & the share of reference options named, lowered by every extra one & Eq.~\ref{eq:metric_choice} \\
list of items & the share of reference items named, lowered by every extra one & Eq.~\ref{eq:metric_choice} \\
numeric & equality at the reference's precision, after normalizing format and units & Eq.~\ref{eq:metric_numeric} \\
open-ended & exact match after normalization, otherwise the token-level F1 with the reference & Eq.~\ref{eq:metric_text} \\
\blockgap
\rowcolor{blue_bg}
\multicolumn{3}{@{}c@{}}{\textcolor{blue_text}{\textbf{\textit{Code Generation}}}} \\
\blockgap
code with unit tests & every test of the problem passes & Eq.~\ref{eq:metric_code} \\
code with a reference checker & the benchmark's own checker accepts the code & Eq.~\ref{eq:metric_code} \\
plotting code & rendered plot accuracy & Eq.~\ref{eq:metric_judge}-\ref{eq:metric_chart} \\
\blockgap
\rowcolor{blue_bg}
\multicolumn{3}{@{}c@{}}{\textcolor{blue_text}{\textbf{\textit{Aggregation}}}} \\
\blockgap
a hardened task & the mean over its sub-questions, an unanswered one scoring zero & Eq.~\ref{eq:metric_task} \\
a dataset & the mean over its test tasks & Eq.~\ref{eq:metric_task} \\
\bottomrule
\end{tabularx}

\vspace{16pt}

\end{table}

\begin{table}[H]

\centering
\caption{\textbf{Data Statistics.} Every dataset \oursdata hardens, the strategies it supports, the task type its answers take, the input its questions carry, and the number of original questions each hardened set holds at level $\ell = 3$ and $\ell = 5$ (written \textit{$\ell=3$ / $\ell=5$}). Counts are questions rather than bundles, since a task carries $\ell$ of them under \textit{Shared} and \textit{Paired}: bundling regroups the same questions into fewer and harder tasks as the level rises, and under \textit{Decoy} the level changes only how many inputs surround the one that answers the question. A level is kept only when its test side holds at least 100 original questions (\S\ref{appendix:subsec:data_adaptation:selection}), which is why two \textit{Shared} levels are absent. A dash "--" in the training column marks a dataset held out for evaluation only, whose questions all serve the test side.}
\label{tab:data_stats}

\vspace{3pt}

\small
\renewcommand{\arraystretch}{1.2}
\newcommand{\blockgap}{\noalign{\vskip 2pt}}
\newcommand{\dsline}{\arrayrulecolor{black!25}\specialrule{0.4pt}{0.45ex}{0.45ex}\arrayrulecolor{black}}

\begin{tabularx}{\linewidth}{@{}p{0.215\linewidth}>{\centering\arraybackslash}p{0.085\linewidth}>{\centering\arraybackslash}p{0.125\linewidth}>{\centering\arraybackslash}p{0.06\linewidth}>{\centering\arraybackslash}p{0.17\linewidth}>{\centering\arraybackslash}X@{}}
\toprule
\multicolumn{1}{@{}c}{\textbf{Dataset}} & \multicolumn{1}{c}{\textbf{Strategy}} & \multicolumn{1}{c}{\textbf{Task Type}} & \multicolumn{1}{c}{\textbf{Input}} & \multicolumn{1}{c}{\textbf{Test Questions}} & \multicolumn{1}{c@{}}{\textbf{Training Questions}} \\
\midrule
\blockgap
\rowcolor{blue_bg}
\multicolumn{6}{@{}c@{}}{\textcolor{blue_text}{\textbf{\textit{Document Understanding}}}} \\
\blockgap
\multirow{3}{*}{\shortstack[l]{MMLongBench-Doc \\ \citep{ma2024mmlongbenchdoc}}} & Shared & mixed & file & 960 / 820 & -- \\
 & Paired & mixed & file & 1,089 / 1,090 & -- \\
 & Decoy & mixed & file & 1,091 / 1,091 & -- \\
\dsline
\multirow{3}{*}{\shortstack[l]{LongDocURL \\ \citep{deng2025longdocurl}}} & Shared & mixed & file & 456 / 405 & 1,443 / 1,160 \\
 & Paired & mixed & file & 540 / 540 & 1,782 / 1,780 \\
 & Decoy & mixed & file & 541 / 541 & 1,783 / 1,783 \\
\blockgap
\rowcolor{blue_bg}
\multicolumn{6}{@{}c@{}}{\textcolor{blue_text}{\textbf{\textit{Finance Understanding}}}} \\
\blockgap
\multirow{3}{*}{\shortstack[l]{DocFinQA \\ \citep{reddy2024docfinqa}}} & Shared & numeric & text & 609 / 275 & 5,031 / 4,250 \\
 & Paired & numeric & text & 843 / 840 & 5,733 / 5,735 \\
 & Decoy & numeric & text & 843 / 843 & 5,735 / 5,735 \\
\blockgap
\rowcolor{blue_bg}
\multicolumn{6}{@{}c@{}}{\textcolor{blue_text}{\textbf{\textit{Multimodal Slide Understanding}}}} \\
\blockgap
\multirow{3}{*}{\shortstack[l]{SlideVQA-Doc \\ \citep{tanaka2023slidevqa}}} & Shared & open-ended & file & 1,791 / 1,390 & 8,637 / 7,230 \\
 & Paired & open-ended & file & 2,214 / 2,215 & 10,617 / 10,615 \\
 & Decoy & open-ended & file & 2,215 / 2,215 & 10,617 / 10,617 \\
\blockgap
\rowcolor{blue_bg}
\multicolumn{6}{@{}c@{}}{\textcolor{blue_text}{\textbf{\textit{Chart Understanding}}}} \\
\blockgap
\multirow{2}{*}{\shortstack[l]{InfoChartQA-Text \\ \citep{xie2025infochartqa}}} & Shared & mixed & image & 2,610 / 2,365 & -- \\
 & Paired & mixed & image & 3,129 / 3,125 & -- \\
\dsline
\multirow{2}{*}{\shortstack[l]{InfoChartQA-Visual \\ \citep{xie2025infochartqa}}} & Shared & mixed & image & 132 / -- & 576 / -- \\
 & Paired & mixed & image & 1,065 / 1,735 & 4,401 / 7,180 \\
\dsline
\multirow{2}{*}{\shortstack[l]{InfoChartQA-Metaphor \\ \citep{xie2025infochartqa}}} & Shared & multiple-choice & image & 309 / -- & -- \\
 & Paired & multiple-choice & image & 333 / 330 & -- \\
\blockgap
\rowcolor{blue_bg}
\multicolumn{6}{@{}c@{}}{\textcolor{blue_text}{\textbf{\textit{Science}}}} \\
\blockgap
\shortstack[l]{MMMU-Pro \\ \citep{yue2025mmmupro}} & Paired & multiple-choice & image & 1,728 / 1,730 & -- \\
\blockgap
\rowcolor{blue_bg}
\multicolumn{6}{@{}c@{}}{\textcolor{blue_text}{\textbf{\textit{Math}}}} \\
\blockgap
\shortstack[l]{We-Math \\ \citep{qiao2024wemath}} & Paired & multiple-choice & image & 1,740 / 1,740 & -- \\
\blockgap
\rowcolor{blue_bg}
\multicolumn{6}{@{}c@{}}{\textcolor{blue_text}{\textbf{\textit{Coding}}}} \\
\blockgap
\shortstack[l]{TACO \\ \citep{li2023taco}} & Paired & code generation & text & 1,362 / 1,360 & 5,424 / 5,425 \\
\dsline
\shortstack[l]{DS-1000 \\ \citep{lai2022ds1000}} & Paired & code generation & text & 873 / 870 & -- \\
\dsline
\shortstack[l]{ChartCoder \\ \citep{zhao2025chartcoder}} & Paired & code generation & image & 999 / 1,000 & -- \\
\bottomrule
\end{tabularx}

\vspace{6pt}

\end{table}

\subsection{Data Statistics}
\label{appendix:subsec:data_adaptation:stats}

Tab.~\ref{tab:data_stats} summarizes all the datasets \oursdata hardens, together with their task types, input types, and the size of each hardened set at both levels. The level shapes these sizes directly: \textit{shared} and \textit{paired} divide a split's questions into tasks of $\ell$ questions, so raising the level yields fewer and harder tasks out of nearly the same pool of questions; while \textit{decoy} keeps one question per task and raises the number of inputs it hides the gold one among, which leaves its task count unchanged. Because a task carries every question in its bundle, the test side of a kept level holds at least 100 original questions to form a valid test set.

\oursdata hardens $12$ datasets across $7$ domains. We only use four datasets to supply training datasets, which are split before hardening, so no hardened test task shares a question with training. We keep the remaining eight datasets to be evaluation-only. The hardened datasets span various types of answers, including numeric, multiple-choice, open-ended, and code, and a single task may mix several types. Input types range from slides and documents to charts and figures.

Each dataset supports only the hardening strategies its source data supports (\S\ref{appendix:subsec:data_adaptation:selection}), so the number of evaluation arms differs across datasets. \oursdata provides $44$ arms for three regimes:

\begin{oursenumerate}
\item \textbf{In-distribution ($4$ arms).} The four datasets \ours trains on, each taken at the \textit{paired} strategy and $\ell{=}3$: InfoChartQA-Visual, LongDocURL, SlideVQA-Doc, and TACO. Training sees no other arm.
\item \textbf{Out-of-distribution ($13$ arms).} The same four datasets under a strategy, a level, or both that training never sees, yielding $5$ arms at $\ell{=}3$ and $8$ arms at $\ell{=}5$.
\item \textbf{Out-of-domain ($27$ arms).} Eight datasets \ours never trains on: InfoChartQA-Text, InfoChartQA-Metaphor, DS-1000, MMLongBench-Doc, DocFinQA, MMMU-Pro, We-Math, and ChartCoder, yielding $14$ arms at $\ell{=}3$ and $13$ arms at $\ell{=}5$.
\end{oursenumerate}

Fig.~\ref{fig:exp_heterogeneity} shows every method on each of these arms.


\section{\ours: Inside the Learning}
\label{appendix:sec:method_extension}

\subsection{Credit the Structure: Traced Credit, Learnable Co-Evolution}
\label{appendix:subsec:method_extension:credit}


\textbf{Only genuine root causes are charged.}
A failed node that has a failed dependency, direct or indirect, is treated as a cascaded failure, so its credit propagates to the root failure along the workflow's edges, for any topology. Infrastructure faults, such as a provider connection error or a content filter, are not the policy's fault and thus excluded. A node is judged by its final occurrence in $\tau_j$, so a step that fails early in a loop but recovers later is not charged. A fault that never surfaces as a node failure is left to the global signal.

\textbf{Per-role credit makes co-evolution learnable.}
For example, when a workflow runs both the \textit{Generator}'s design and the \textit{Inventor}'s components, one scalar entangles their contributions. Localized credit restricts each role's credit term to its own nodes, so a role whose footprint is narrower than the workflow is not charged for failures outside it, \textit{e.g.}, the \textit{Inventor} is not charged for failures on nodes that use none of the components it creates, as the roles co-evolve (\S\ref{subsec:method:evolve}). This local credit is a bounded term of the global hierarchy in \S\ref{subsec:method:reward}, alongside the role-specific validity $v^\rho$ that measures each role's own contribution (Tab.~\ref{tab:role_validity})

\subsection{Harness the Workflow: Rich in Signal, Broad in Reach}
\label{appendix:subsec:method_extension:reward}


\textbf{Every step of progress earns signal.}
The ladder makes the signal rich: it reflects every step of a role's progress, from a well-formed output through a valid contribution to a correct answer. Because each stage with a nonzero weight adds its own term, the policy learns from the full hierarchy of outcomes. Thus, even a partially successful workflow yields a graded, informative gradient.

\textbf{One shared harness generalizes to every role.}
The invariant tail $w_e\, e + w_a\, a$ carries across every role: the harness always runs a workflow and evaluates its answer, and only the \textit{source} of the workflow or of its execution differs. Every role's own quality enters through its own \textit{validity} reward $v^\rho$ at the same stage of the ladder, deterministic and never a learned judge. Because the whole framework is shared (Fig.~\ref{fig:upstream_downstream_structure}), the harness generalizes to any role, which is what makes co-evolution across arbitrary roles practical (\S\ref{subsec:method:evolve}).

\subsection{Define the Reward: Validity Reward in the Hierarchical Ladder}
\label{appendix:subsec:method_extension:validity}

The reward of Eq.~\ref{eq:hierarchical_reward} is built from five terms. Whether the role's output \textit{parses} ($f$) gates the rest: an output that cannot be read leaves nothing to execute. Whether the role's contribution is \textit{valid} ($v^\rho$) is a role-specific term, and \S\ref{appendix:subsec:method_extension:reward} states why the remaining terms are not: the harness always runs a workflow and grades its answer, so whether the workflow \textit{executes} to a final answer ($e$) and how \textit{correct} that answer is ($a$) are shared by every role (Eq.~\ref{eq:hierarchical_reward}). The credit term $c_j^\rho$ (\S\ref{appendix:subsec:method_extension:credit}) then charges a role only for the nodes it authors. Thus, only a parse failure gates the reward, while validity rewards without gating. A correct answer requires an execution that reaches one, so a role earns dense partial credit for how far its contribution carries the workflow.

A role earns its validity reward when its own output effectively contributes to the workflow, and Tab.~\ref{tab:role_validity} shows what a contribution amounts to for each role. Four forms arise: (1) \textit{structural correctness} rewards a workflow that is well formed against the pool; (2) \textit{deciding and authoring} reward components that resolve to an existing pool entry or, when newly authored, instantiate and run; (3) \textit{helpfulness} rewards what the role's intervention adds, calculated as the graded gain over the same flow without it, suiting a role that steers the flow; and (4) \textit{step format} and \textit{step liveness} reward every node the role runs for reaching an answer of its own. A reward resting on a semantic judgment (\textit{e.g.}, the coherence of a plan) is deliberately excluded: a semantically correct output can also lead to the wrong final answer, whereas an outcome-grounded gain measures the actual contribution.


\begin{table}[t]

\centering
\caption{\textbf{Per-Role Validity.} What each role's own output needs to satisfy, split into the upstream roles that build a workflow and the downstream agent that executes it (Fig.~\ref{fig:upstream_downstream_structure}). Every reward is deterministic, so no learned judge enters the reward (\S\ref{appendix:subsec:method_extension:reward}).}
\label{tab:role_validity}

\small

\renewcommand{\arraystretch}{1.2}
\newcommand{\blockgap}{\noalign{\vskip 2pt}}

\begin{tabularx}{\linewidth}{@{}p{0.215\linewidth}p{0.265\linewidth}X@{}}
\toprule
\multicolumn{1}{@{}c}{\textbf{Role}} & \multicolumn{1}{c}{\textbf{Format Reward ($f$)}} & \multicolumn{1}{c@{}}{\textbf{Validity Reward ($v^\rho$)}} \\
\midrule
\rowcolor{blue_bg}
\multicolumn{3}{@{}c@{}}{\textcolor{blue_text}{\textbf{\textit{Upstream: Building a Workflow}}}} \\
\blockgap
\textit{Planner} & its plan parses & \textit{helpfulness}: the graded gain its plan brings over the same flow without one \\
\textit{Generator} & its workflow parses & \textit{structure}: the workflow is a valid graph over the pool(s) \\
\textit{Inventor} \newline (\textit{Generator Skill}) & every creation decision and authored component parses & \textit{deciding} and \textit{authoring}: reused components resolve to the pool, and newly authored ones instantiate and run \\
\blockgap
\rowcolor{blue_bg}
\multicolumn{3}{@{}c@{}}{\textcolor{blue_text}{\textbf{\textit{Downstream: Executing a Workflow}}}} \\
\blockgap
\textit{Downstream} agent & its output parses & \textit{step format} and \textit{step liveness}: every node it runs reaches an answer of its own \\
\bottomrule
\end{tabularx}

\vspace{6pt}

\end{table}

\subsection{Wire the Workflow: Representing Workflows In Different Topologies}
\label{appendix:subsec:method_extension:topologies}

\begin{figure}[t]
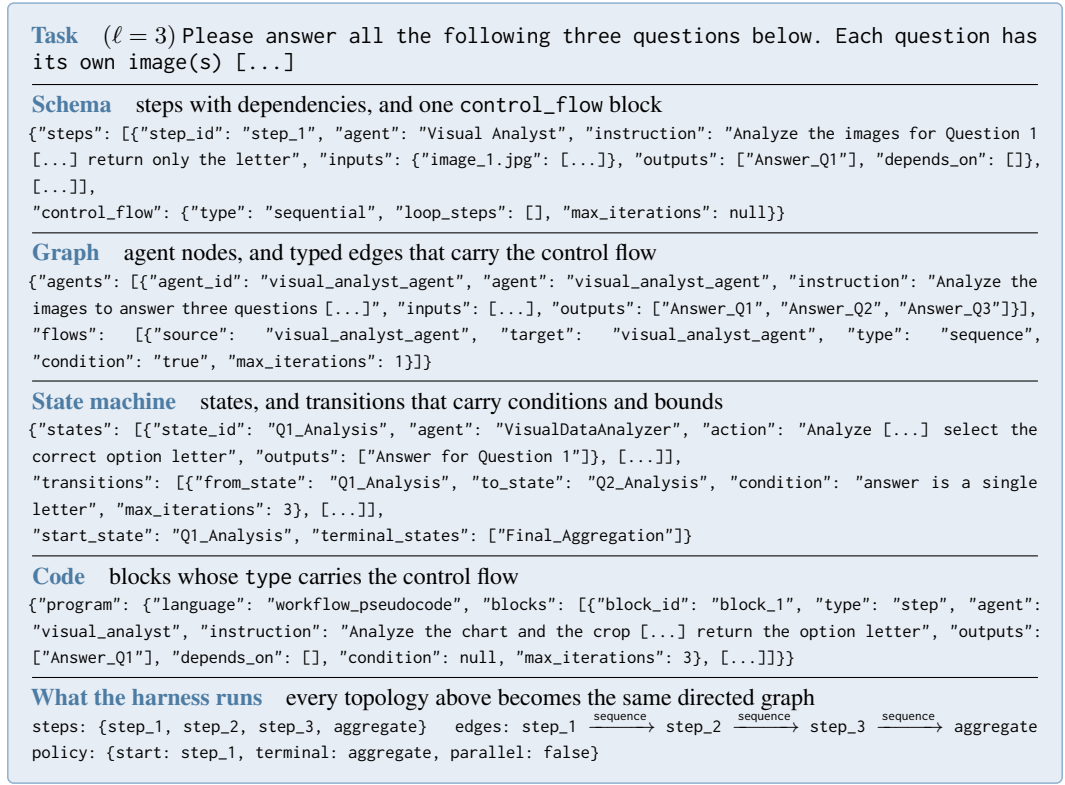

\centering
\small
\begin{tcolorbox}[colback=bboxface, colframe=bboxedge, boxrule=0.6pt, arc=2pt, left=6pt, right=6pt, top=5pt, bottom=5pt]

{\bfseries\textcolor{bboxtag}{Task}}\quad $(\ell=3)$ \texttt{Please answer all the following three questions below. Each question has its own image(s) \texttt{[...]}}

\vspace{4pt}\hrule\vspace{4pt}

{\bfseries\textcolor{bboxtag}{Schema}}\quad steps with dependencies, and one \texttt{control\_flow} block\\
{\ttfamily\scriptsize \{"steps": [\{"step\_id": "step\_1", "agent": "Visual Analyst", "instruction": "Analyze the images for Question 1 \texttt{[...]} return only the letter", "inputs": \{"image\_1.jpg": \texttt{[...]}\}, "outputs": ["Answer\_Q1"], "depends\_on": []\}, \texttt{[...]}],\\
"control\_flow": \{"type": "sequential", "loop\_steps": [], "max\_iterations": null\}\}}

\vspace{4pt}\hrule\vspace{4pt}

{\bfseries\textcolor{bboxtag}{Graph}}\quad agent nodes, and typed edges that carry the control flow\\
{\ttfamily\scriptsize \{"agents": [\{"agent\_id": "visual\_analyst\_agent", "agent": "visual\_analyst\_agent", "instruction": "Analyze the images to answer three questions \texttt{[...]}", "inputs": \texttt{[...]}, "outputs": ["Answer\_Q1", "Answer\_Q2", "Answer\_Q3"]\}],\\
"flows": [\{"source": "visual\_analyst\_agent", "target": "visual\_analyst\_agent", "type": "sequence", "condition": "true", "max\_iterations": 1\}]\}}

\vspace{4pt}\hrule\vspace{4pt}

{\bfseries\textcolor{bboxtag}{State machine}}\quad states, and transitions that carry conditions and bounds\\
{\ttfamily\scriptsize \{"states": [\{"state\_id": "Q1\_Analysis", "agent": "VisualDataAnalyzer", "action": "Analyze \texttt{[...]} select the correct option letter", "outputs": ["Answer for Question 1"]\}, \texttt{[...]}],\\
"transitions": [\{"from\_state": "Q1\_Analysis", "to\_state": "Q2\_Analysis", "condition": "answer is a single letter", "max\_iterations": 3\}, \texttt{[...]}],\\
"start\_state": "Q1\_Analysis", "terminal\_states": ["Final\_Aggregation"]\}}

\vspace{4pt}\hrule\vspace{4pt}

{\bfseries\textcolor{bboxtag}{Code}}\quad blocks whose \texttt{type} carries the control flow\\
{\ttfamily\scriptsize \{"program": \{"language": "workflow\_pseudocode", "blocks": [\{"block\_id": "block\_1", "type": "step", "agent": "visual\_analyst", "instruction": "Analyze the chart and the crop \texttt{[...]} return the option letter", "outputs": ["Answer\_Q1"], "depends\_on": [], "condition": null, "max\_iterations": 3\}, \texttt{[...]}]\}\}}

\vspace{4pt}\hrule\vspace{4pt}

{\bfseries\textcolor{bboxtag}{What the harness runs}}\quad every topology above becomes the same directed graph\\
{\ttfamily\scriptsize steps: \{step\_1, step\_2, step\_3, aggregate\} \quad edges: step\_1 $\xrightarrow{\text{sequence}}$ step\_2 $\xrightarrow{\text{sequence}}$ step\_3 $\xrightarrow{\text{sequence}}$ aggregate \quad policy: \{start: step\_1, terminal: aggregate, parallel: false\}}

\end{tcolorbox}
\caption{\textbf{Workflow Topologies.} We show an example of the same hardened task represented in four topologies: a step schema, an agent graph, a state machine, and a block code. Each topology example is directly extracted from our evaluation outputs and trimmed with \texttt{[...]}. The four topologies differ only in where the control flow is written down, \textit{e.g.}, a separate \texttt{control\_flow} block, a typed edge, a transition, a block type, \textit{etc}. Each is then converted into a shared directed graph structure, so execution, credit, and evaluation stay uniform without knowing which topology produced it (\S\ref{appendix:subsec:method_extension:topologies}).}
\label{fig:appendix:topologies}
\end{figure}

A workflow needs to be properly written down to enable seamless upstream generation and downstream execution. \ours allows four workflow topologies (Fig.~\ref{fig:appendix:topologies}): a step schema, an agent graph, a state machine, and a block code.
%
%
The four topologies differ only in where the control flow is recorded. A schema states it in a separate block, a graph types each edge, a state machine attaches conditions and bounds to transitions, and a code carries it in the type of a block. Each is converted into the same directed graph before harness execution: nodes hold the agent with its instruction and inputs, and edges hold sequence, parallel, conditional and bounded-loop relations, and a cycle that no topology marks as a loop is reclassified as a bounded one so the graph stays schedulable. The control operations convert the same way in every topology. Execution, credit, and evaluation therefore read one graph and never observe which topology produces it. For this reason, \ours wires different topologies into the shared workflow graph (\S\ref{subsec:method:wire}), and thus the topology a generator writes in cannot change what the harness measures.

What the choice does change is how reliably the generator writes a workflow worth running: whether its output parses, whether the structure it declares is valid, and how well the resulting workflow performs. Thus, this becomes an empirical question that we measure via an ablation over the four topologies in \S\ref{subsec:results:ablation}.

\section{Implementation Details}
\label{appendix:sec:implementation}

\subsection{Reward Configuration}
\label{appendix:subsec:implementation:weights}

Every role instantiates one hierarchical ladder, so the reward weights differ only where the roles differ. We summarize per-role reward weights in Tab.~\ref{tab:reward_weights}. Format carries $0.1$ for every role, since parsing is a gate rather than an achievement. Across different roles, the answer term always dominates at $0.6$ to $0.8$. We set the execution weight to $0.0$ for every role in main experiments: a workflow that does not run cannot answer, so the answer term already covers it, and weighting execution on its own would pay a workflow for running cleanly while answering wrongly. Coding tasks reward the term at $0.1$, since a workflow there reaches an answer only once its code executes, so the ladder sets this weight by task type (\S\ref{appendix:subsec:results:exec_reward}). The validity term holds what each role alone controls, and its components take the weight the answer term leaves: one component for the \textit{Generator} and the steering roles, two for the \textit{Inventor} and the \textit{Downstream} agent. The credit penalty of \S\ref{appendix:subsec:method_extension:credit} rides on top at $0.1$, and the ablation that switches it off sets it to $0.0$ (\S\ref{subsec:results:ablation}).

\begin{table}[t]

\centering
\caption{\textbf{Per-Role Reward Configuration.} Every role instantiates the same hierarchical ladder (Eq.~\ref{eq:hierarchical_reward}). The validity term is role-specific, and the credit penalty rides on top of the weighted sum (\S\ref{appendix:subsec:method_extension:credit}).}
\label{tab:reward_weights}

\small

\renewcommand{\arraystretch}{1.2}

\begin{tabularx}{\linewidth}{@{}p{0.20\linewidth}cXcc@{}}
\toprule
\multicolumn{1}{@{}c}{\textbf{Role}} & \textbf{Format} ($w_f$) & \multicolumn{1}{c}{\textbf{Validity} ($w_v$)} & \textbf{Answer} ($w_a$) & \textbf{Credit} ($w_c$) \\
\midrule
\textit{Planner}    & $0.1$ & \textit{helpfulness} $0.1$                                         & $0.8$ & $0.1$ \\
\textit{Generator}  & $0.1$ & \textit{structure} $0.1$                                           & $0.8$ & $0.1$ \\
\textit{Inventor} \newline (\textit{Generator Skill}) & $0.1$ & \textit{grounding} $0.1$ \; + \; \textit{authoring} $0.1$ & $0.7$ & $0.1$ \\
\textit{Critic}     & $0.1$ & \textit{helpfulness} $0.3$                                         & $0.6$ & $0.1$ \\
\textit{Downstream} agent & $0.1$ & \textit{step format} $0.025$ \; + \; \textit{step liveness} $0.075$ & $0.8$ & $0.1$ \\
\bottomrule
\end{tabularx}

\vspace{6pt}

\end{table}

\subsection{Experiment Configuration}
\label{appendix:subsec:implementation:config}

As \ours contributes a training paradigm rather than an optimizer, the objective of Eq.~\ref{eq:objective} admits any policy-gradient method. We instantiate it with GRPO, DAPO, and CISPO, ablating the choice. All three draw $m$ rollouts per task and score them against their group rather than a learned critic, so they differ in how that group is turned into an update: GRPO normalizes token-level advantages within the group, DAPO decouples the clipping bounds and drops the groups that carry no signal, and CISPO clips the importance-sampling weight instead of the update, so a token far from the old policy is down-weighted rather than dropped from the gradient.
We summarize our experiment configuration details in Tab.~\ref{tab:exp_config}.

\begin{table}[t]

\centering
\caption{\textbf{Experiment Configuration.} We summarize the experiment configurations for training and evaluation (\S\ref{sec:exps}). The same configuration applies across all four evolution modes and roles.}
\label{tab:exp_config}

\vspace{3pt}

\small

\renewcommand{\arraystretch}{1.2}
\newcommand{\blockgap}{\noalign{\vskip 2pt}}

\begin{tabularx}{\linewidth}{@{}p{0.24\linewidth}>{\centering\arraybackslash}p{0.28\linewidth}X@{}}
\toprule
\multicolumn{1}{@{}c}{\textbf{Setting}} & \textbf{Value} & \multicolumn{1}{c@{}}{\textbf{Note}} \\
\midrule
\rowcolor{blue_bg}
\multicolumn{3}{@{}c@{}}{\textcolor{blue_text}{\textbf{\textit{Training Configuration}}}} \\
\blockgap
Policy optimization   & GRPO / DAPO / CISPO & ablated; all three instantiate Eq.~\ref{eq:objective} \\
Learning rate         & $1\times10^{-6}$ & constant, uniform for all settings \\
Group size $m$        & $8$ & rollouts per task \\
Batch size            & $8$ & tasks per step \\
Epochs per data pass  & $1$ & constant, uniform for all settings \\
Clipping ratio        & $0.2$ & symmetric for GRPO; DAPO decouples the upper bound \\
KL loss coefficient   & $0.001$ & against the reference policy \\
Rollout sampling      & $T{=}1.0$, top-$p{=}0.95$, top-$k{=}20$ & temperature kept at $1.0$ for group diversity \\
\blockgap
\rowcolor{blue_bg}
\multicolumn{3}{@{}c@{}}{\textcolor{blue_text}{\textbf{\textit{Evaluation Configuration}}}} \\
\blockgap
Decoding              & greedy, $T{=}0$ & deterministic, so a reported number is reproducible \\
Baseline compute budget       & matched & each baseline receives the budget of the workflow it is compared against \\
Seed                  & $12345$ & shared by hardening, splitting, and evaluation \\
\bottomrule
\end{tabularx}

\vspace{6pt}

\end{table}

\subsection{Notation}
\label{appendix:subsec:implementation:notation}

We summarize the core notations in Tab.~\ref{tab:notation}. An index $i$ denotes the $i$-th item of the list in scope, and in \S\ref{sec:method} it indexes tasks.

\begin{table}[h]

\centering
\caption{\textbf{Notation.} We summarize notations used in our paper, grouped by the concept they denote, from tasks and workflows (\S\ref{sec:prelim}) to rollouts, credit, reward, optimization (\S\ref{sec:method}), and data hardening (\S\ref{appendix:sec:data_adaptation}). In \S\ref{sec:method}, the indices $i$, $j$, and $k$ run over tasks, rollouts, and distinct workflows, respectively, and symbols indexed by $\rho$ are specific to role $\rho$, \textit{e.g.}, $c_j^\rho$ is the credit of role $\rho$ in the $j$-th rollout.}
\label{tab:notation}
\small
\renewcommand{\arraystretch}{1.2}
\newcommand{\blockgap}{\noalign{\vskip 2pt}}

\begin{tabularx}{\linewidth}{@{}p{0.22\linewidth}Xp{0.10\linewidth}@{}}
\toprule
\multicolumn{1}{@{}c}{\textbf{Symbol}} & \multicolumn{1}{c}{\textbf{Meaning}} & \multicolumn{1}{c@{}}{\textbf{Introduced in}} \\

\midrule

\rowcolor{blue_bg}
\multicolumn{3}{@{}c@{}}{\textcolor{blue_text}{\textbf{\textit{Tasks and Workflows}}}} \\
\blockgap
$x \sim \mathcal{X}$, $x_i$ & a task drawn from the task distribution; $x_i$ is the $i$-th task & \S\ref{sec:prelim}, \ref{sec:method} \\
$G = (V, E)$ & a workflow: a directed graph whose nodes $V$ are agents, tools, or skills, and whose edges $E$ carry the data and control flow & \S\ref{sec:prelim} \\
$\mathcal{G}$ & the space of workflows & \S\ref{sec:method} \\
$y$ & the answer a workflow returns & Eq.~\ref{eq:pipeline} \\
\blockgap
\rowcolor{blue_bg}
\multicolumn{3}{@{}c@{}}{\textcolor{blue_text}{\textbf{\textit{Harness and Signal}}}} \\
\blockgap
$\mathcal{H}$ & the harness, which executes and grades a workflow & Eq.~\ref{eq:pipeline} \\
$h(G, x) \in [0, 1]$ & the harness signal, serving as the evaluation metric, the training signal, and the test-time objective & Eq.~\ref{eq:pipeline} \\
$\tau_j$ & the execution trace of $G_j$ produced by $\mathcal{H}$ & \S\ref{subsec:method:credit} \\
\blockgap
\rowcolor{blue_bg}
\multicolumn{3}{@{}c@{}}{\textcolor{blue_text}{\textbf{\textit{Roles and Policies}}}} \\
\blockgap
$\rho \in \mathcal{R}$ & a role, and the set of roles & \S\ref{sec:prelim} \\
$\pi_\rho$, $\theta_\rho$ & the policy of role $\rho$, and its parameters & \S\ref{sec:prelim} \\
$\pi_g$ & the policy of the \textit{Generator} & Eq.~\ref{eq:pipeline} \\
$\rho'$ & ranges over all roles in the flow & Eq.~\ref{eq:objective} \\
\blockgap
\rowcolor{blue_bg}
\multicolumn{3}{@{}c@{}}{\textcolor{blue_text}{\textbf{\textit{Pool}}}} \\
\blockgap
$\Omega$ & the pool of reusable components a workflow is built from & \S\ref{sec:prelim} \\
$\Omega_{\textit{agent}}$, $\Omega_{\textit{tool}}$, $\Omega_{\textit{skill}}$ & the agent, tool, and skill pools; $\Omega$ is their union under modular granularity & \S\ref{subsec:method:wire} \\
$\Omega_{\textit{capsule}}$ & the capsule pool; $\Omega = \Omega_{\textit{capsule}}$ under coarse granularity & \S\ref{subsec:method:wire} \\
\blockgap
\rowcolor{blue_bg}
\multicolumn{3}{@{}c@{}}{\textcolor{blue_text}{\textbf{\textit{Rollouts}}}} \\
\blockgap
$j$, $m$ & the rollout index, and the number of rollouts of a role being optimized & Fig.~\ref{fig:flowright_main} \\
$k$, $M$ & the distinct-workflow index, and the number of distinct workflows the $m$ rollouts span & Fig.~\ref{fig:flowright_main} \\
$G_j = (V_j, E_j)$ & the workflow the $j$-th rollout is graded on & Fig.~\ref{fig:flowright_main} \\
$o_j^\rho$ & the output of role $\rho$ in rollout $j$ & \S\ref{subsec:method:reward} \\
$O^\rho = \{o_j^\rho\}_{j=1}^{m}$ & the $m$ rollouts of role $\rho$ being optimized & Fig.~\ref{fig:flowright_main} \\
\blockgap
\rowcolor{blue_bg}
\multicolumn{3}{@{}c@{}}{\textcolor{blue_text}{\textbf{\textit{Credit and Reward}}}} \\
\blockgap
$V_j^\rho \subseteq V_j$ & the nodes of $G_j$ that role $\rho$ authors & \S\ref{subsec:method:credit} \\
$F_j^\rho \subseteq V_j^\rho$ & the nodes of role $\rho$ that root-cause a failure & \S\ref{subsec:method:credit} \\
$c_j^\rho \in [-1, 0]$ & the structure-aware credit of role $\rho$ & Eq.~\ref{eq:credit} \\
$f$, $v^\rho$, $e$, $a$ & whether the output parses, whether role $\rho$'s contribution is valid, whether the workflow executes to a final answer, and how correct the answer is & \S\ref{subsec:method:reward} \\
$w_f, w_v, w_e, w_a, w_c$ & per-role weighting coefficients & Eq.~\ref{eq:hierarchical_reward} \\
$h_j^\rho = h^\rho(G_j, x_i)$ & the hierarchical reward of role $\rho$ in rollout $j$ & Eq.~\ref{eq:hierarchical_reward} \\
\blockgap

\rowcolor{blue_bg}
\multicolumn{3}{@{}c@{}}{\textcolor{blue_text}{\textbf{\textit{Optimization}}}} \\
\blockgap
$\mathcal{J}(\pi_\rho)$ & the objective optimized for role $\rho$ & Eq.~\ref{eq:objective} \\
$P$, $P^\star$ & a reusable prior that conditions generation, and the optimal prior & Eq.~\ref{eq:distill} \\
\blockgap

\bottomrule

\end{tabularx}

\vspace{6pt}

\end{table}


\begin{table}[h]

\vspace{-20pt}

\centering
\small

\renewcommand{\arraystretch}{1.2}
\newcommand{\blockgap}{\noalign{\vskip 2pt}}

\begin{tabularx}{\linewidth}{@{}p{0.22\linewidth}Xp{0.10\linewidth}@{}}
\toprule
\multicolumn{1}{@{}c}{\textbf{Symbol}} & \multicolumn{1}{c}{\textbf{Meaning}} & \multicolumn{1}{c@{}}{\textbf{Introduced in}} \\

\midrule

\rowcolor{blue_bg}
\multicolumn{3}{@{}c@{}}{\textcolor{blue_text}{\textbf{\textit{Data Hardening}}}} \\
\blockgap
$\mathcal{B}$, $(q, z, y^\star)$ & a source dataset, and one of its samples: a question $q$, its input $z$, and its reference answer $y^\star$ & Alg.~\ref{alg:data_adaptation} \\
$\sigma$ & a hardening strategy: \textit{Shared}, \textit{Paired}, or \textit{Decoy} & \S\ref{appendix:subsec:data_adaptation:strategies} \\
$\ell$ & the hardening level: questions per task for \textit{Shared} and \textit{Paired}, and inputs per task for \textit{Decoy} & \S\ref{appendix:subsec:data_adaptation:strategies} \\
$\mathcal{T}^{\mathrm{train}}_{\sigma}$, $\mathcal{T}^{\mathrm{test}}_{\sigma}$ & the hardened tasks that strategy $\sigma$ builds from a dataset's training split, and those it builds from its test split & Alg.~\ref{alg:data_adaptation} \\
$|x|$ & the number of sub-questions a hardened task $x$ holds, one per original sample bundled into it & Alg.~\ref{alg:data_adaptation} \\
\blockgap
\rowcolor{blue_bg}
\multicolumn{3}{@{}c@{}}{\textcolor{blue_text}{\textbf{\textit{Evaluation Metrics}}}} \\
\blockgap
$y_i$, $y^\star_i$ & the answer a workflow returns for the $i$-th sub-question of a hardened task, and that sub-question's reference answer & Eq.~\ref{eq:metric_task} \\
$s(y_i, y^\star_i) \in [0, 1]$ & the score of one sub-question, given by the metric that its answer type uses & Tab.~\ref{tab:metrics} \\
$s_{\mathrm{set}}$, $s_{\mathrm{num}}$, $s_{\mathrm{text}}$, $s_{\mathrm{code}}$ & the metrics of set-valued, numeric, open-ended, and code answers & Eq.~\ref{eq:metric_choice}--\ref{eq:metric_code} \\
$S(x)$ & a hardened task's score: the mean of the scores of its sub-questions & Eq.~\ref{eq:metric_task} \\
$\mathrm{Acc}$ & a dataset's accuracy: the mean task score over the test tasks of a strategy and level & Eq.~\ref{eq:metric_task} \\
$Y$, $Y^\star$ & the options, or list items, that an answer names, and those that its reference names & Eq.~\ref{eq:metric_choice} \\
$\mathrm{round}_{y^\star}(\cdot)$ & rounding to the precision that the reference answer states & Eq.~\ref{eq:metric_numeric} \\
$\nu(\cdot)$ & normalization of case, punctuation, and spacing in a text answer & Eq.~\ref{eq:metric_text} \\
$\mathrm{tok}(\cdot)$ & the multiset of tokens of a text answer & Eq.~\ref{eq:metric_text} \\
$\one$ & the indicator: 1 when its condition holds, and 0 otherwise & Eq.~\ref{eq:metric_numeric} \\
\textsc{Check} & the strictest execution-based test that a code dataset provides for generated code & Eq.~\ref{eq:metric_code} \\
\textsc{Judge} & an LLM judge's verdict: one integer on the anchored scale its prompt fixes & Eq.~\ref{eq:metric_judge} \\
$L$ & the top of that anchored scale, by which the verdict is normalized & Eq.~\ref{eq:metric_judge} \\
$s_{\mathrm{judge}}$ & the metric of an answer no rule decides: a judge's verdict over $L$ & Eq.~\ref{eq:metric_judge} \\
\textsc{Exec} & whether generated code executes successfully, beyond \textsc{Check} assessment above & Eq.~\ref{eq:metric_chart} \\
\textsc{Render} & the chart image that executing the code produces & Eq.~\ref{eq:metric_chart} \\
$s^{\mathrm{code}}_{\mathrm{chart}}$, $s^{\mathrm{render}}_{\mathrm{chart}}$, $s_{\mathrm{chart}}$ & the code and render judges of a chart answer, and their mean & Eq.~\ref{eq:metric_chart} \\

\bottomrule

\end{tabularx}

\end{table}

\section{A Deeper Dive Beneath the Results}
\label{appendix:sec:results}

\begin{figure}[h]
    \centering
    \includegraphics[width=1.0\textwidth]{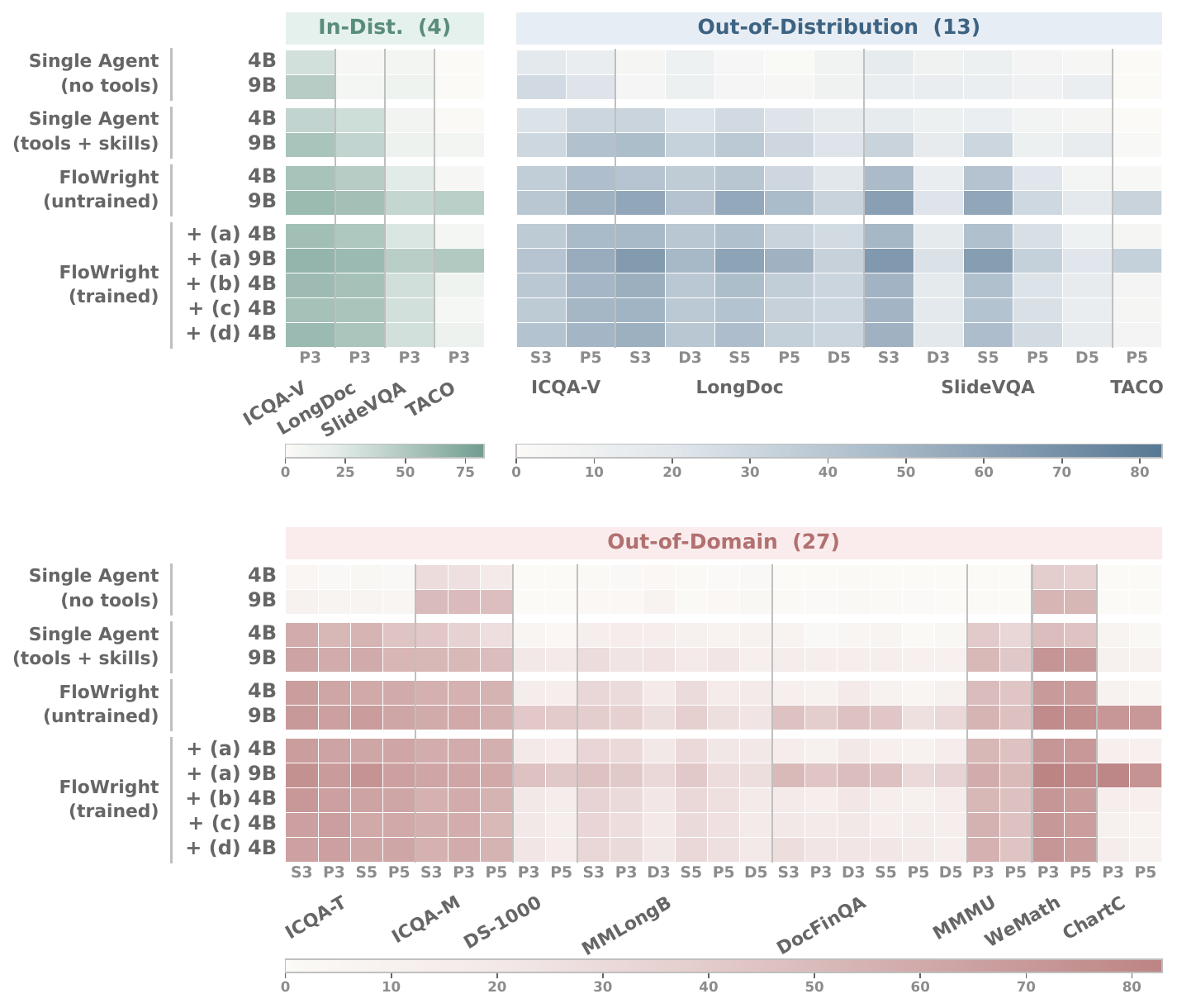}
    
    \caption{\textbf{Per-Arm Results Behind the Mean Scores.} Complementing the mean accuracy scores in Tab.~\ref{tab:exp_main}, we show the accuracy (\%) of each method on the $44$ evaluation arms, grouped into the three regimes (Tab.~\ref{tab:exp_main}, \S\ref{appendix:subsec:data_adaptation:stats}): \colorbox{tea_bg}{\textcolor{tea_text}{\textbf{in-distribution}}}, \colorbox{blue_bg}{\textcolor{blue_text}{\textbf{out-of-distribution}}} and \colorbox{rose_bg}{\textcolor{rose_text}{\textbf{out-of-domain}}}. We denote the hardening strategy and level as: S/P/D = Shared/Paired/Decoy, 3/5 = level. Color intensity encodes accuracy on a shared scale, and hue identifies the regime.}
    \label{fig:exp_heterogeneity}
\end{figure}

\subsection{Unfold the Means: Consistent Gains Across Datasets, Strategies, and Levels}
\label{appendix:subsec:results:heterogeneity}

\textbf{\ours improves performance across diverse datasets, hardening strategies, and task complexities.}
As shown in Fig.~\ref{fig:exp_heterogeneity}, the improvement holds on every dataset: under multi-agent co-evolution, all $12$ datasets gain over the untrained workflow, with up to $+7.34\%$ on LongDocURL and $+5.13\%$ on ChartCoder. So no domain is traded away for another. It also holds on every hardening strategy, and grows as the strategy demands more of the workflow: $+3.57\%$ on \textit{shared}, $+4.11\%$ on \textit{paired}, and $+4.68\%$ on \textit{decoy}. The improvement further holds at both levels of task complexity, with multi-agent co-evolution gaining $+4.31\%$ out-of-domain at $\ell{=}3$ and $+4.09\%$ at $\ell{=}5$, and $+5.80\%$ and $+6.21\%$ out-of-distribution (Tab.~\ref{tab:exp_main}). Consequently, increased task complexities does not erode the gain. Harnessing the workflow therefore improves performance across different datasets, strategies, and levels \oursdata spans, rather than on a favorable subset of them.

\subsection{Generalize the Harness: Optimization Transfers to Different Settings}
\label{appendix:subsec:results:generalization}

\begin{figure}[h]
    \centering
    \includegraphics[width=1.0\textwidth]{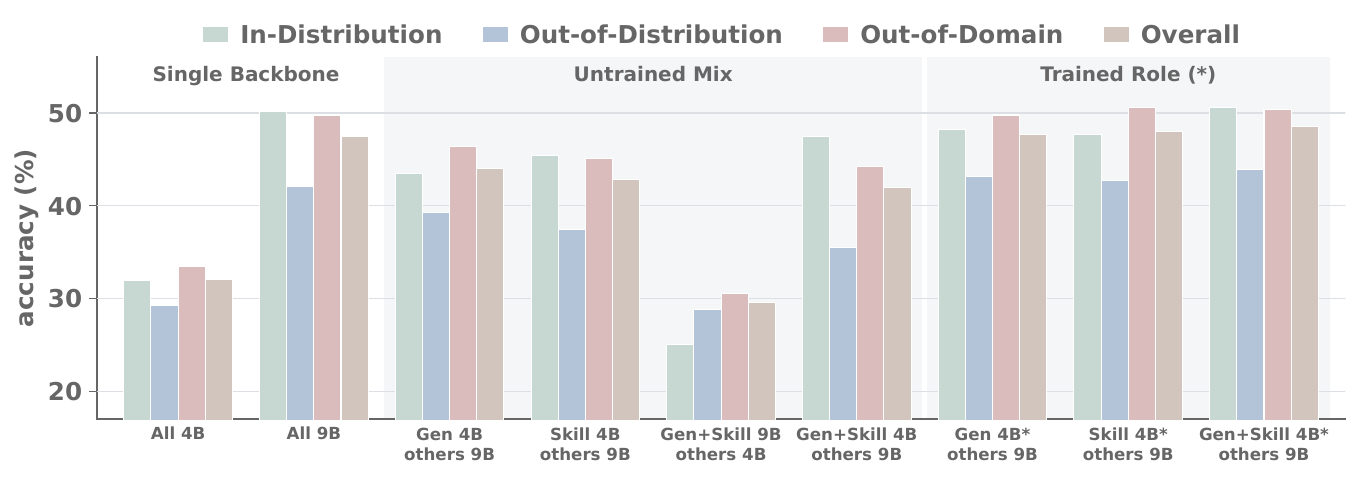}

    \vspace{-10pt}

    \caption{\textbf{Roles Powered by Different Backbones.} Accuracy (\%) across three evaluation regimes over all $44$ arms (\S\ref{sec:exps}). \textit{Single Backbone} uses one backbone for every role. \textit{Untrained Mix} uses a \texttt{4B} or \texttt{9B} backbone to specific roles, with no role optimized. \textit{Trained Role} ($*$) uses the RL-trained \texttt{4B} role, with \texttt{Qwen3.5-9B} powering other roles. \textit{Gen} and \textit{Skill} name the \textit{Generator} and the skill it co-evolves with, and \textit{others} covers every remaining role.}
    \label{fig:exp_mix_model}
\end{figure}

\textbf{\ours generalizes to roles powered by backbones it never trains alongside.}
As shown in Fig.~\ref{fig:exp_mix_model}, powering one upstream role with \texttt{Qwen3.5-4B} and other roles with \texttt{Qwen3.5-9B} performs between the two single-agent baselines, gaining $+11.91\%$ over the all-\texttt{4B} setting when \texttt{Qwen3.5-4B} powers the \textit{Generator} and $+10.77\%$ when \texttt{Qwen3.5-4B} powers the \textit{Inventor} skill. Exchanging the two models reverses the result: powering the \textit{Generator} and the \textit{Inventor} skill with \texttt{Qwen3.5-9B} and the downstream agent with \texttt{Qwen3.5-4B} reaches $29.57\%$, below the all-\texttt{4B} baseline of $32.09\%$. A stronger upstream model thus builds workflows for a downstream agent of its own capability, which a less capable downstream agent executes less effectively. Consequently, when using different backbones to power upstream and downstream roles, the downstream agent bounds what a workflow performs however well it is built. Optimizing a single \texttt{4B} role then raises accuracy above the all-\texttt{9B} baseline of $47.50\%$, to $47.67\%$ for the \textit{Generator}, $47.96\%$ for the \textit{Inventor} skill, and $48.50\%$ for co-evolving the two together. \textit{An optimized role transfers its gain when other roles run on a different backbone it never trains with.} Train-time optimization therefore generalizes across models that power the remaining roles.

\begin{figure}[h]
    \centering
    \includegraphics[width=1.0\textwidth]{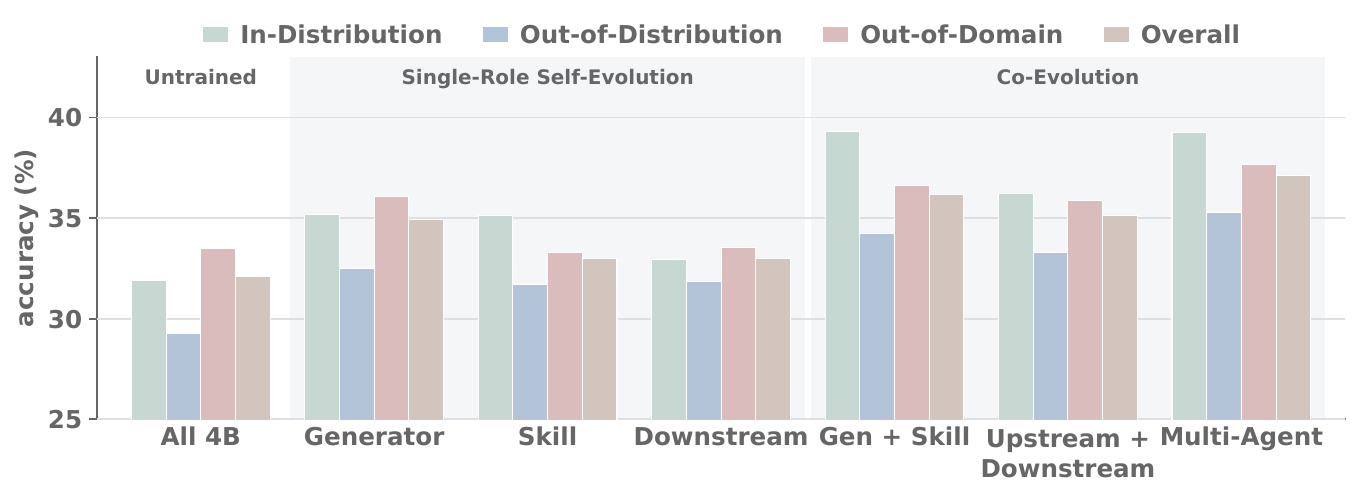}

    \vspace{-12pt}

    \caption{\textbf{Evolution of Different Roles.} Accuracy (\%) across three evaluation regimes over all $44$ arms (\S\ref{sec:exps}). Every role runs \texttt{Qwen3.5-4B}, with the only variable as which role the harness optimizes. \textit{Single-Role Self-Evolution} optimizes the \textit{Generator}, skill, and downstream agent individually. \textit{Co-Evolution} optimizes two or more together (\S\ref{subsec:method:evolve}).}
    \label{fig:exp_train_roles}
\end{figure}

\textbf{\ours improves performance whichever role it optimizes.}
As shown in Fig.~\ref{fig:exp_train_roles}, every single-role optimization outperforms the untrained baseline, and every co-evolving optimization gains more: $+2.83\%$ for the \textit{Generator}, $+0.90\%$ for the \textit{Inventor} skill, and $+0.88\%$ for the downstream agent alone, against $+4.06\%$ for \textit{Generator}-skill co-evolution, $+3.06\%$ for upstream-downstream co-evolution, and $+5.03\%$ for multi-agent co-evolution. The \textit{Generator} is the largest single lever, gaining $+2.59\%$ out-of-domain where the other two roles stay flat. \textit{The role that designs every workflow returns the largest gain among single roles.}

\clearpage

\begin{figure}[h]
    
    \centering
    \includegraphics[width=1.0\textwidth]{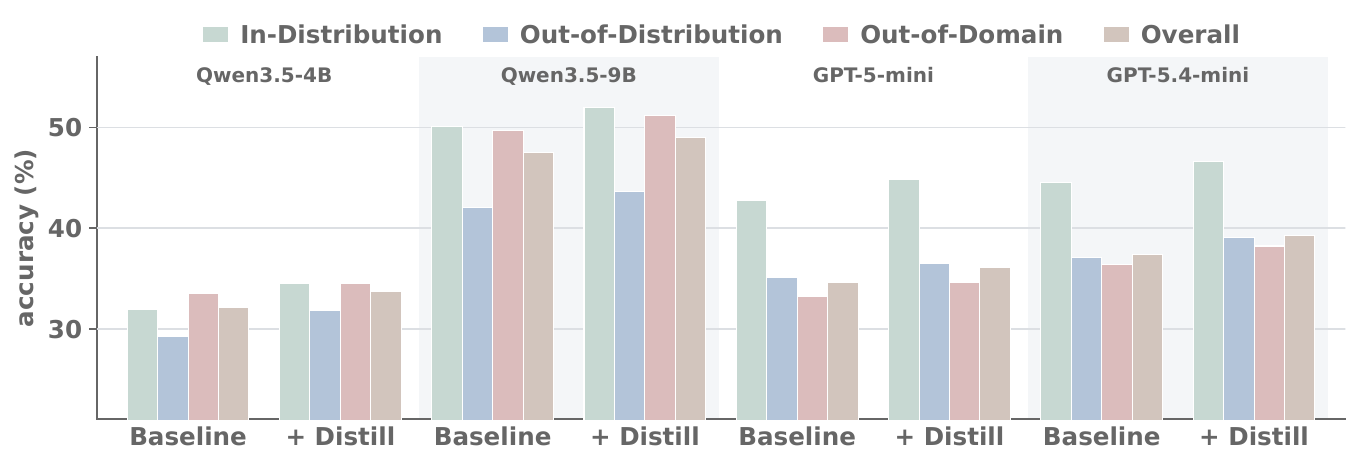}

    \vspace{-10pt}

    \caption{\textbf{Meta Distillation Across Backbones.} Accuracy (\%) across three evaluation regimes over all $44$ arms (\S\ref{sec:exps}). Each backbone appears as an untrained \textit{Baseline}, compared with \textit{+Distill} providing the meta prior distilled at five shots (\S\ref{subsec:method:objective}).}
    \label{fig:exp_gpt_models}

    \vspace{-12pt}
\end{figure}

\clearpage

\textbf{Meta distillation improves every backbone, including those \ours never optimizes.}
As shown in Fig.~\ref{fig:exp_gpt_models}, a meta-distilled prior gains in all three regimes on all four backbones, and overall by $+1.62\%$ on \texttt{Qwen3.5-4B}, $+1.53\%$ on \texttt{Qwen3.5-9B}, $+1.44\%$ on \texttt{GPT-5-mini}, and $+1.91\%$ on \texttt{GPT-5.4-mini}. Every backbone keeps its weights fixed, so each gain comes from the prior alone, and \texttt{GPT-5.4-mini} gains most among the four. \textit{What the harness distills at test time transfers to models whose weights stay beyond reach.} Optimizing the signal thus improves a model \ours trains and a model it only prompts.

\subsection{Weigh the Execution: The Ladder Adapts to the Task Type}
\label{appendix:subsec:results:exec_reward}

\begin{wrapfigure}[20]{r}{0.5\textwidth}
    \vspace{-12pt}
    \centering
    \includegraphics[width=0.5\textwidth]{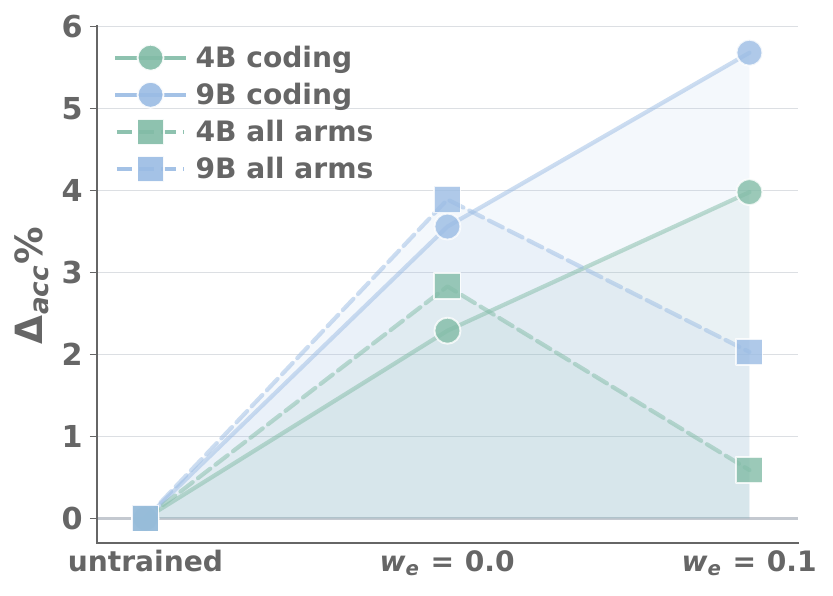}

    \vspace{-10pt}

    \caption{\textbf{Execution Reward by Task Type.} $\Delta_\textit{acc}$(\%) against untrained baselines, on the four coding arms (TACO and DS-1000) and over all $44$ arms (\S\ref{sec:exps}). Every other ladder weight stays fixed (Tab.~\ref{tab:reward_weights}).}
    \label{fig:exp_exec_reward}

\end{wrapfigure}

\textbf{Weighting execution adds the most on coding tasks.}
As shown in Fig.~\ref{fig:exp_exec_reward}, an execution weight of $w_e{=}0.1$ improves over the untrained baselines, gaining $+3.98\%$ and $+5.68\%$ on the coding arms and $+0.59\%$ and $+2.03\%$ over all $44$ arms, for \texttt{Qwen3.5-4B} and \texttt{Qwen3.5-9B} respectively. On coding, where a workflow reaches an answer only once its code executes to earn score based on the portion of evaluation tests it passes, $w_e{=}0.1$ outperforms $w_e{=}0.0$ by $+1.69\%$ and $+2.12\%$. Over all arms, where the answer term already covers whether a workflow reaches an answer, $w_e{=}0.0$ leads by $+2.24\%$ and $+1.86\%$. This is also why our main experiments hold the execution weight at $0.0$ (\S\ref{appendix:subsec:implementation:weights}). \textit{The hierarchical ladder therefore servers as a paradigm whose weights follow what each task type demands, with execution earning its own weight where the task rewards running the workflow.}

\end{document}